\documentclass[conference]{IEEEtran}
\IEEEoverridecommandlockouts
\usepackage{cite}
\usepackage{amsmath,amssymb,amsfonts}
\usepackage{algorithmic}
\usepackage{algorithm}
\usepackage{graphicx}
\usepackage[dvipsnames]{xcolor}
\usepackage{textcomp}
\usepackage{subfigure}
\usepackage{xcolor}
\usepackage{fancyhdr}

\usepackage{booktabs,ragged2e}
\usepackage[flushleft]{threeparttable}

\makeatletter
\renewcommand\IEEEkeywordsname{Keywords}
\def\BibTeX{{\rm B\kern-.05em{\sc i\kern-.025em b}\kern-.08em
    T\kern-.1667em\lower.7ex\hbox{E}\kern-.125emX}}
\begin{document}
\title{Robust Smart Home Face Recognition under Starving Federated Data
% \thanks{*Corresponding author}
% \thanks{978-1-6654-7477-1/22/\$31.00 ©2022 IEEE}
% \copyrightnotice{978-1-6654-7477-1/22/\$31.00 ©2022 IEEE}
}

\author{\IEEEauthorblockN{Jaechul Roh} 
\IEEEauthorblockA{\textit{Dept. of Electronic and Computer Engineering}\\
\textit{HKUST}\\
Hong Kong, Hong Kong \\
jroh@connect.ust.hk} \\%

\and
\IEEEauthorblockN{Yajun Fang\IEEEauthorrefmark{1}}
\thanks{\IEEEauthorrefmark{1}Corresponding author}
\IEEEauthorblockA{\textit{Universal Village Society} \\
1 Broadway, Cambridge, MA, 02142 \\
yjfang@mit.edu} \\
}

% \IEEEoverridecommandlockouts
% \IEEEpubid{\makebox[\columnwidth]{978-1-6654-7477-1/22/\$31.00 ©2022 IEEE
% \hfill} \hspace{\columnsep}\makebox[\columnwidth]{ }}

\maketitle

\IEEEpubidadjcol

\begin{abstract}

Over the past few years, the field of adversarial attack received numerous attention from various researchers with the help of successful attack success rate against well-known deep neural networks that were acknowledged to achieve high classification ability in various tasks. However, majority of the experiments were completed under a single model, which we believe it may not be an ideal case in a real-life situation. In this paper, we introduce a novel federated adversarial training method for smart home face recognition, named FLATS, where we observed some interesting findings that may not be easily noticed in a traditional adversarial attack to federated learning experiments. By applying different variations to the hyperparameters, we have spotted that our method can make the global model to be robust given a starving federated environment. Our code can be found on \verb|https://github.com/jcroh0508/FLATS|. 

\end{abstract}

% Introduction
\section{Introduction}
The introduction of Deep Neural Networks (DNNs) to the field of machine learning grasped the attention of numerous researchers by achieving the classification ability to almost perfection where no other system was able to achieve. With the help of huge data storage capacity and swift data transmission speed, proposed neural network based architectures became highly applicable in various areas such as medicine, finance, art and even in law. However, most models that reached the state-of-the-art (SOTA) results were based on clean data, which some research has started to prove that they can be extremely vulnerable and less secure to poisoned data. Szegedy \cite{szegedy2013intriguing} first introduced the concept of adversarial attack where a slight perturbation to any given input data can lead to a misclassification of DNNs. Although such proposed method was slow and less robust compared to recently presented adversarial attack methods \cite{goodfellow2014explaining, wong2020fast, madry2017towards, andriushchenko2020square}, it became the offset of a new paradigm in the Artificial Intelligence (AI) security and privacy research.  

\begin{figure}[htp]
    \centering
    \includegraphics[width=265pt]{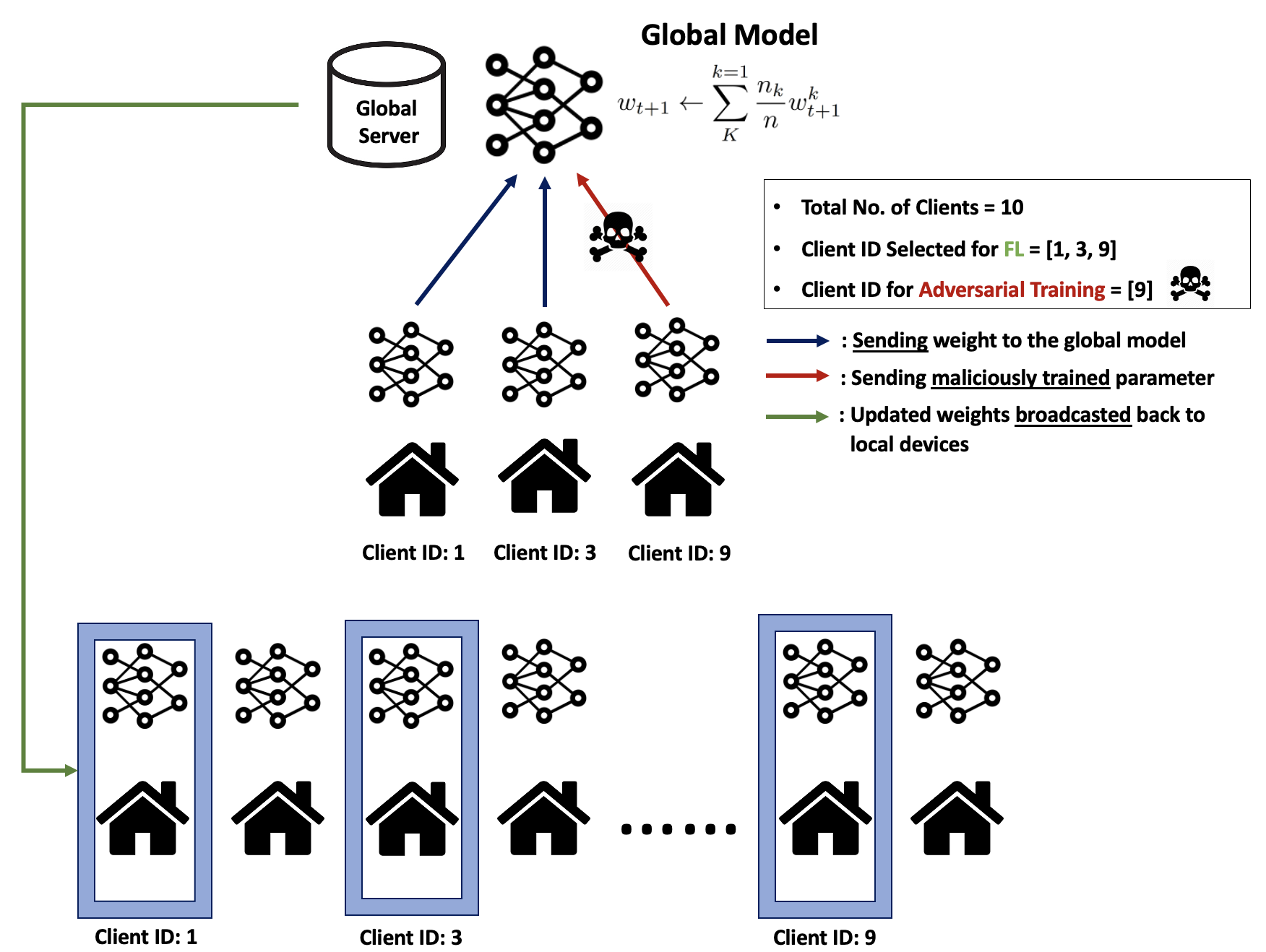}
    \caption{General Architecture of FLATS}
    \label{fig:Robust_FL}
\end{figure}

The field of robustness built much reputation when algorithms such as \textit{Fast Gradient Sign Method (FGSM)} \cite{goodfellow2014explaining} became well-known for being swift in terms of creating adversarial examples \cite{goodfellow2014explaining}. However, models that went through adversarial training with FGSM \cite{goodfellow2014explaining} were neutralized by various new attack methods such as \textit{Projected Gradient Descent (PGD)} \cite{madry2017towards} and \textit{C\&W Attack} \cite{carlini2017towards}. Moreover, "adversarially" trained DNNs with PGD was not only robust against the adversarial examples created with PGD \cite{madry2017towards} itself, but also against C\&W and FGSM \cite{goodfellow2014explaining} attacks, which amplified the capability of transferable adversarial examples \cite{tramer2017space, madry2017towards}. The field later discovered stronger and more robust attack algorithms that could effectively process adversarial training to the models. But, the problem aroused when adversarial attack is applied in a federated learning \cite{mcmahan2017communication} based system. 

With the help of federated learning \cite{mcmahan2017communication}, numerous companies were able to train large neural network models in a more computationally inexpensive way while preserving privacy for the individual devices at the same time. Since only the model parameter, rather than the data, is sent to the global server for aggregation, individual users or even institutions do no have to worry about their private data being exposed. However, global model accuracy can be degraded when malicious users include adversarial examples within the local model training process, which disturbs the global parameter of the aggregated model. In addition, the model will always be vulnerable to the perturbed data if it never handled any adversarial training. Standard training will not be sufficient enough for them to be robust against the attacks. For such reasons, in depth research in the field of combination between adversarial attack and federated learning is crucial. 

In this paper, we propose a novel federated adversarial training method, named FLATS (\textbf{F}ederated \textbf{L}earning \textbf{A}dversarial \textbf{T}raining for \textbf{S}mart Home Face Recognition System). Our method surprisingly helps the global model performance against various adversarial attacks in both white-box and black-box settings with a limited amount of distributed data, which we refer to as "starving" federated data or "starving" federated environment. FLATS can be directly utilized in a real-life application where it aids to preserve the robustness of the global model against different poisonous attacks, while each devices are only allocated with a starving data. 

Moreover, by adding various data manipulation to specific devices, our method improves both the global and the robust accuracy of the aggregated model. After further evaluating on the augmented test data that consists of face images with different brightness, the model tends to maintain its classification ability on both the brightness manipulated data and even against various adversarial examples. Such experiment could indirectly reveal the issue of fairness that exists within the neural network and our empirical results may demonstrate on reducing such bias in a "shallow" level. We certainly believe that our method may not just be limited within the scope of robust machine learning but it is also capable of broadening the spectrum towards alleviating the well-known concerns that needs to be resolved in the general computer vision and face recognition tasks. 

The paper is structured in the following manner. It first introduces the concept of federated learning \cite{mcmahan2017communication} and few  adversarial attack methods that will be used in our investigation (Section II). Then, it illustrates our proposed robust federated learning algorithms and how they will be applied according to different experimental settings (Section III). Furthermore, it demonstrates the empirical results of the conducted experiments depending on the modifications we have added with help of tables, graphical representations, and the analysis of such results for a better understanding (Section IV). Section V and Section VI concludes the paper by exemplifying future studies and indicating the limitations of our proposed method as well as the experiment.

% Related Works
\section{Related Works}

\subsection{Face Recognition in Smart Home}
Face recognition became an essential factor in terms of improving the general security of the smart home system. Munir et al.\cite{munir2019face} and Zuo et al.\cite{zuo2005real} has focused on the software technology aspect of the face recognition architecture typically used for the smart home system, while Kak et al.\cite{kak2019smart} and Pawar et al.\cite{pawar2018smart} suggested the general structure of the interaction between hardware devices and software face recognition model used in smart home. 

\subsection{Fast Gradient Sign Method (FGSM)}
Given the model \(C(x) = y\) with an input data of \textit{x} and the target label \textit{y}, the goal of adversarial attack is to create an adversarial example (\(x^*\)) that results in \(C(x^*) \neq y\). The attack can be either targeted or untargeted depending on the attacker.  Fast Gradient Sign Method, also known as FGSM \cite{goodfellow2014explaining}, is considered as one of the fastest algorithm to create an adversarial example to induce misclassification of the deep learning models. The equation to create the adversarial example is as follows:

\begin{IEEEeqnarray}{rCl}
 x^* = x + \epsilon sign(\nabla _xJ(\theta, x, y)) \;.
\label{eq:fgsm_eq}
\end{IEEEeqnarray}

According to formula (1), to create an adversarial example (\(x^*\)) we add the original input data "\textit{x}" with the perturbation rate (\(\epsilon\)) multiplied by the sign of the loss between "\textit{x}" and the true label \textit{y}. Since FGSM \cite{goodfellow2014explaining} uses \(L_\infty\) norm, the pixels will be modified in every dimension by the value of \(\epsilon\). In order to defend against such attack, the paper introduces a new adversarial objective function \cite{goodfellow2014explaining} as the following:

\begin{IEEEeqnarray}{rCl}
\widetilde{J} = \alpha J(\theta, x, y) + (1-\alpha)J(\theta, x + \epsilon(\nabla _xJ(\theta, x, y)) \;.
\label{eq:fgsm_loss}
\end{IEEEeqnarray}

We may notice that the objective function is consisted of two parts. \(J(\theta, x, y)\) is for the original input data \textit{x} and \(J(\theta, x + \epsilon(\nabla _xJ(\theta, x, y))\) is the cost function for the adversarial example. \(\alpha\) decides the ratio of considering the cost of adversarial examples when calculating the total loss. 

\subsection{Fast adversarial training using FGSM (FFGSM)}
Although FGSM\cite{goodfellow2014explaining} is considered as the fastest method to create an adversarial example, FGSM-based adversarial training\cite{madry2017towards} is proven to be weak and less robust against other adverasrial attacks such as PGD\cite{madry2017towards}, C\&W\cite{carlini2017towards}, etc. 

Wong et al.\cite{wong2020fast} has simply added a random initialization point \cite{wong2020fast} to the original FGSM \cite{goodfellow2014explaining} adversarial training algorithm. The paper has argued that initiating the adversarial training process from a non-zero perturbation \cite{wong2020fast} is the key to creating a model that is robust against PGD attack while maintaining a similar speed as FGSM \cite{goodfellow2014explaining} based adversarial training. 

% Square Attack  
\subsection{Square Attack}
\textit{Square Attack} \cite{andriushchenko2020square} is a score-based black-box attack algorithm that is utilized in \(l_2\)- and \(l_\infty\)-norm distance measure, which achieved higher attack success rate compared to other black-box attack methods. Since such attack does not depend on the gradient  of the local model, it is not influenced by gradient masking \cite{papernot2017practical}, which refers to the case when the defender tries to hide the gradient information from the attacker. The algorithm processes random search to find the optimal parameter, \(\delta\), and terminates as soon as adversarial example (\(\hat{x}\)) is found, which emphasizes the advantage of query-efficient. 

\subsection{Federated Learning}
Federated learning \cite{mcmahan2017communication} is a concept of training decentralized devices locally and updating the weight of the aggregated model. One of the main advantages of such method is that the local data will neither be sent to the global server nor it will be shared amongst other devices, which preserves and ensures data privacy. Another advantage of federated learning is the hardware utility. Training a neural network with a large amount of data is computationally expensive as well as time consuming. Such issue becomes more serious if the global server has to combine all the local data together. However, if we train them locally and only send the updated weights for aggregation, the training time will be significantly be reduced. 

Two main settings exist for federated learning: \textbf{IID} (Independent and Identically Distributed) setting and \textbf{Non-IID} setting. IID setting is when the distribution are equal among local devices. The data distribution of the local devices is also the representation of distribution for all devices. On the other hand, for Non-IID setting, the distribution of the data has their own unique characteristics depending on the local devices, which is a more realistic situation. The data size for each different local devices are considered as different. The test accuracy of the aggregated model, also known as the global model, has been proven from \cite{mcmahan2017communication} with 99\% for CNN with 100 clients on the MNIST dataset on a Non-IID setting. 

There are numerous ways to update the weights of the aggregated global model. One of the main method introduced by McMahan\cite{mcmahan2017communication} is named \verb|FederatedAveraging| (\verb |FedAvg|) \cite{mcmahan2017communication}. Equation for updating the global model parameter for the \verb |FedAvg| \cite{mcmahan2017communication} algorithm is as follows: 

\begin{IEEEeqnarray}{rCl}
w_{t+1} \leftarrow \sum_{K}^{k=1} \frac{n_k}{n}w_{t+1}^{k} \;.
\label{eq:fedavg}
\end{IEEEeqnarray}

, where local client, \textit{k}, is the index of the total number of \textit{K} clients, \textit{n} is the data size, \(w_{t+1}^{k}\) is the new weight of the \(k^{th}\) client. Updating the local client is implemented based on normal training with backpropagation and gradient descent \cite{mcmahan2017communication}. Such update will be processed over specified number of rounds. The modified aggregated weight will be sent back to the global model for evaluation on the test data. 

% Algorithm Explanation
\section{Our Approach}
To carefully examine the robustness of federated learning of the face recognition model, we have experimented with two separate methods of applying the adversarial training \cite{madry2017towards}. Although our proposed algorithm may seem similar to those introduced earlier, our main novelty comes from how the passive preparation for the malicious attacks could remarkably aid the neural network to maintain its robustness, under a starving federated environment. In addition, we have added a new hyperparameter, named \textit{Adversarial Training Batch Ratio} (\(ABR\)), which decides the proportion of the local training batch in which the model will face the perturbed data when a specific client is chosen to process adversarial training. 

% Algorithm 1
\subsection{FLATS (Method I)}
For \textit{Method I}, we choose the number of clients to process adversarial training \cite{madry2017towards} after randomly selecting the client IDs to go through federated learning. Such method guarantees that adversarial training \cite{madry2017towards} will be carried out every round. Also, a single client may encounter both the clean data as well as the adversarial examples depending on the randomly chosen index. The steps are shown in Algorithm 1. 

\begin{algorithm}
 \caption{FLATS (Method I)}
 \begin{algorithmic}[1]
\STATE $N = $ Total global rounds \\
\STATE $J = $ Total no. of clients \\
\STATE $d = $ Total data size \\
\STATE $d_j = $ Data size of client $j$ \\
\STATE $n = $ No. of clients selected every round \\
\STATE $n_a = $ No. of clients to go through adversarial training \\
\STATE $w_g = $ Global model parameter \\
\STATE $Clients \gets [w_1, w_2, w_3, ..., w_k] $ \\
\STATE $RoundClients\gets [] $ \\
\STATE $AdvClients \gets [] $ \\

\texttt{\\}
\FOR {$N$} 
    \STATE $UpdatedWeights \gets [] $ \\
    \STATE $RoundClients\gets$ Random($Clients$, $n$) \\
    \STATE $AdvClients\gets$ Random($Round Clients$, $n_a$) \\
    \FOR {$i \gets RoundClients$}
        \IF {$i$ is in $AdvClients$} 
            \STATE $newW \gets$ AdvTraining($Clients$[i]) \\
            \STATE $UpdatedWeights \gets newW $ \\
        
        \ELSE
            \STATE $newW \gets$ ClientUpdate($i$, $Clients$[i]) \\
            \STATE {$UpdatedWeights$ $\gets$ $newW$ } \\
        \ENDIF
    \ENDFOR
    \STATE {$w_g$ $\gets$ \verb|FedAvg|(\(UpdatedWeights\))} \\
\ENDFOR

\end{algorithmic} 
\end{algorithm}

%%%%%%%%%%%%%%%%%%%%%%%%%%%%%%%%%%%%%%%%%%%%
\begin{figure*}[h]
    \centering
    \subfigure[]{\includegraphics[width=0.24\textwidth]{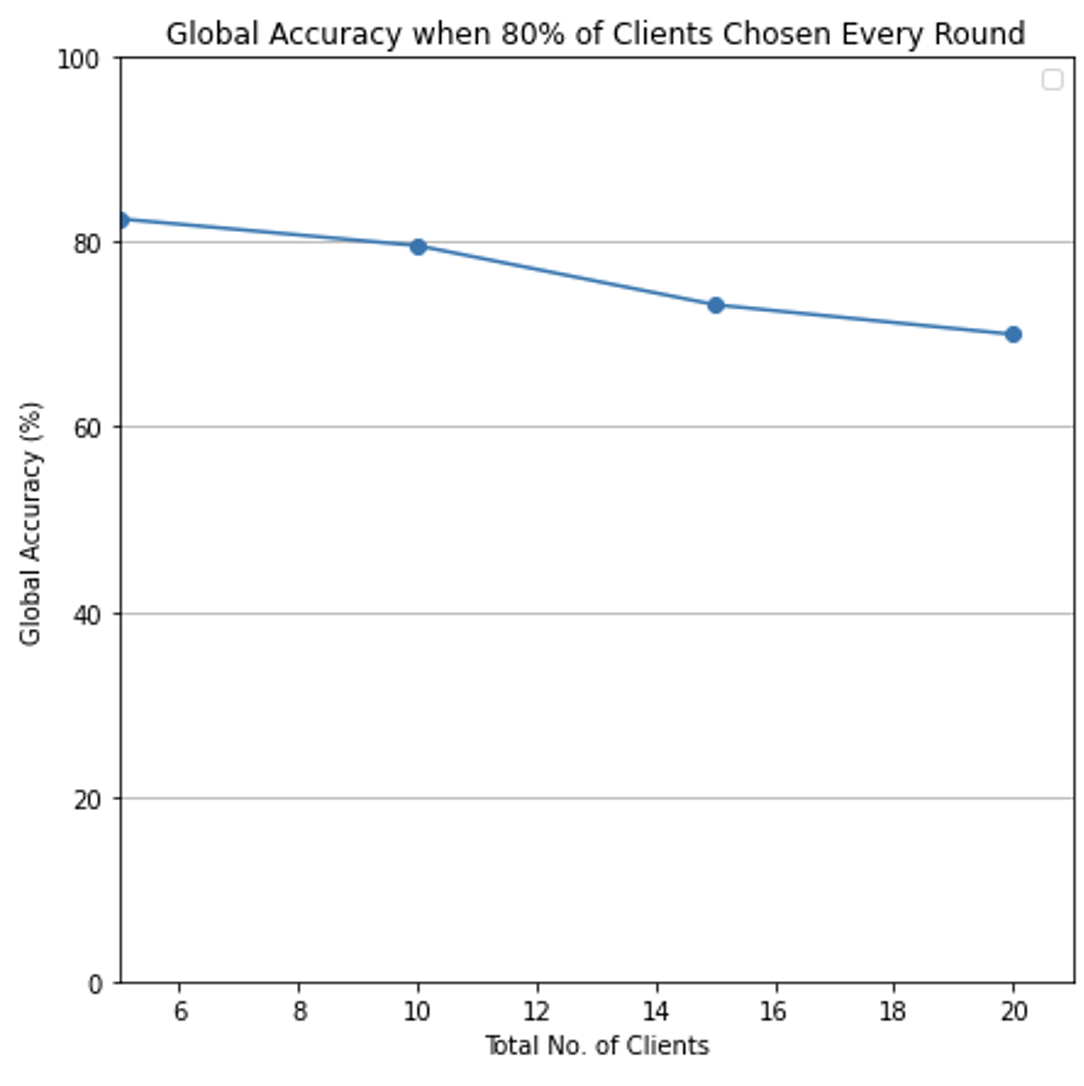}} 
    \subfigure[]{\includegraphics[width=0.24\textwidth]{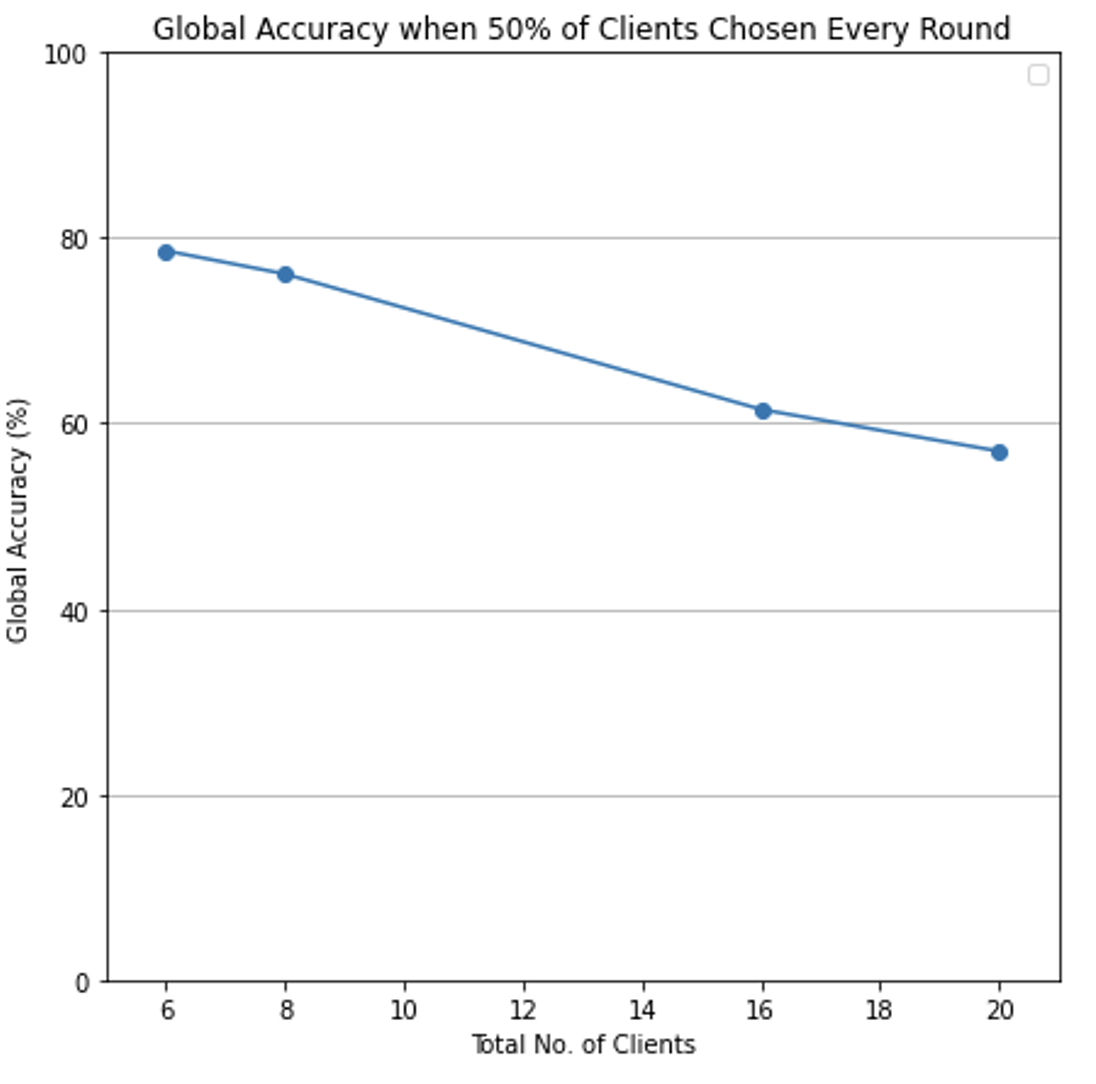}} 
    \subfigure[]{\includegraphics[width=0.24\textwidth]{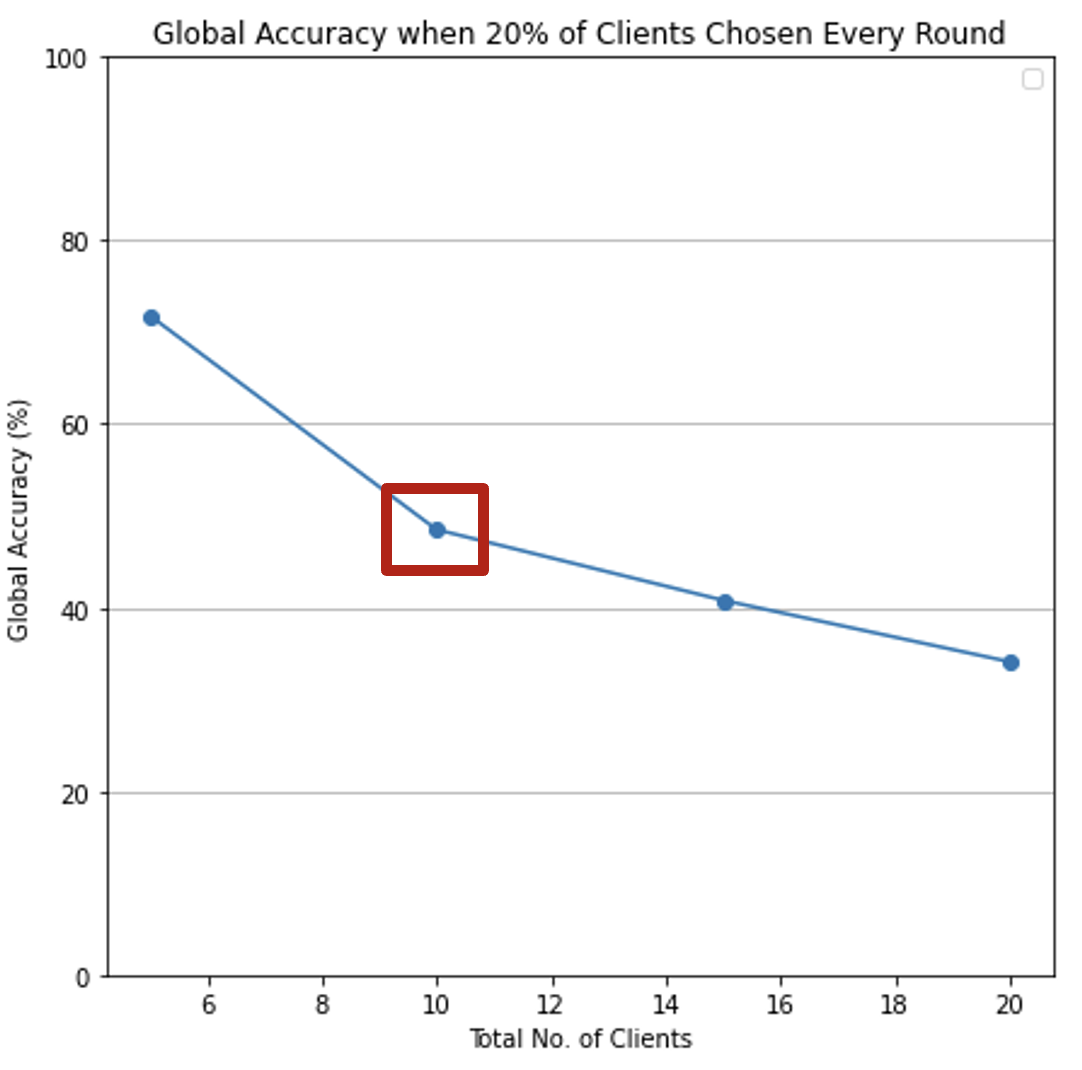}} 
    \caption{Global Acc.(\%) trend depending on proportion of \(n\) chosen (a) 80\% (b) 50\% (c) 20\%}
    \label{fig:foobar}
\end{figure*}
%%%%%%%%%%%%%%%%%%%%%%%%%%%%%%%%%%%%%%%%%%%%

% Algorithm 2
\subsection{FLATS (Method II)}
\textit{Method II} chooses the client IDs for adversarial training \cite{madry2017towards} in the very beginning of the experiment. It is a representation where the randomly chosen clients only consist of adversarial examples, while rest of the devices will be trained with just clean face images. In addition, \textit{Method II} does not guarantee that there will be an adversarial training \cite{goodfellow2014explaining} in each round. The detailed steps are illustrated in Algorithm 2.  

% Algorithm 2
\begin{algorithm}
 \caption{FLATS (Method II)}
  \begin{algorithmic}[1]
\STATE $N = $ Total global rounds \\
\STATE $J = $ Total no. of clients \\
\STATE $d = $ Total data size \\
\STATE $d_j = $ Data size of client $j$ \\
\STATE $n = $ No. of clients selected every round \\
\STATE $n_a = $ No. of clients to go through adversarial training \\
\STATE $w_g = $ Global model parameter \\
\STATE $Clients \gets [w_1, w_2, w_3, ..., w_k] $ \\
\STATE $RoundClients\gets [] $ \\
\STATE $AdvClients \gets [] $ \\

\texttt{\\}
\STATE $AdvClients\gets$ Random($Round Clients$, $n_a$) \\
\FOR {$N$} 
    \STATE $UpdatedWeights \gets [] $ \\
    \STATE $RoundClients\gets$ Random($Clients$, $n$) \\
    \FOR {$i \gets RoundClients$}
        \IF {$i$ is in $AdvClients$} 
            \STATE $newW$ $\gets$ AdvTraining($Clients$[i]) \\
            \STATE $UpdatedWeights$ $\gets$ $newW$ \\
        
        \ELSE
            \STATE $newW \gets$ ClientUpdate($i$, $Clients$[i]) \\
            \STATE $UpdatedWeights$ $\gets$ $newW$ \\
        \ENDIF
    \ENDFOR
    \STATE $w_g$ $\gets$ \verb|FedAvg|(\(UpdatedWeights\)) \\
\ENDFOR

\end{algorithmic} 
\end{algorithm}

\subsection{Pixel / Brightness Modification Algorithm}
We have decided to use \textit{Method I} to further continue with our experiment due to higher global and robust accuracies compared to those of \textit{Method II}. It is easily predictable to anticipate such a result since the algorithm will less often process adversarial training for the global model. Since our purpose is to find a more robust federated learning algorithm for smart home face recognition, we believed that \textit{Method I} is the most suitable choice.

For our next step, we have added an additional condition to our data to observe how global model reacts to such uncommon face data: pixel / brightness modification, and covering the "eye" area of the face image. Such method can be considered as a backdoor attack \cite{gu2017badnets} since these modifications and noises can react as triggers to hinder the classification ability of the global model. 

\section{Experiment}
In this section, we elaborate our experimental setting, specific conditions added for further robustness, and the results depending on such conditions. We also evaluate the results extracted with detailed analysis. 

\subsection{Experimental Setting}
We use the celebrity face dataset from the Kaggle competition that consists of 105 classes \cite{pinsface}. Although there are only a total of 17,534 images, we would like to emphasize the result from our experiment can be a representation of robust starving federated environment. In the real world, we cannot guarantee that each devices will consist of enough face images to help the classification model to reach its maximum accuracy.

First, we preprocess our face images to \(224 \times 224 \times 3\) for a fair experiment. However, we downgrade the pixels for later modification as mentioned in Section III. CosFace \cite{wang2018cosface} and ArcFace \cite{deng2019arcface} are known to be the state-of-the-art (SOTA) deep face recognition models in MegaFace Challenge \cite{kemelmacher2016megaface}, Youtube Faces (YTF) and Labeled Face in the Wild (LFW \cite{srivastava2019performance}), which are gigantic face image datasets. For example, MegaFace \cite{kemelmacher2016megaface} consists of 1M face images with 690K unique identifications \cite{deng2019arcface}. However, we have used \textbf{ResNet-34} \cite{he2016deep} and \textbf{105-class celebrity face dataset} \cite{pinsface} that achieved \textbf{97.8\% }test accuracy under normal deep learning training method. 

We have decided to not utilize the SOTA architectures as well as the representative face recognition benchmarks in our experiment because they can be computationally expensive and significantly time consuming in terms of training time. The whole process will take even longer if we apply them to our robust federated learning algorithm. Some  CosFace \cite{wang2018cosface} and ArcFace \cite{deng2019arcface} uses ResNet \cite{he2016deep} as their baseline model. Since we are working with comparatively smaller face dataset, we believed that directly applying ResNet-34 \cite{he2016deep} to our investigation is sufficient enough. 

For our entire experiment, we have used a batch size of 64 for the training data and 32 for the test data. Stochastic Gradient Descent (SGD) with a learning rate of 0.5 was used in stead of Adam for the optimizer since it achieves higher global accuracy for Non-IID federated learning compared to that used with Adam. In addition, for creating the adversarial examples, we have employed the Torchattacks \cite{kim2020torchattacks} PyTorch repository to accelerate our experimental procedure. 

% Benign Federated Learning
\subsection{Benign Federated Learning (IID)}

%%%%%%%%%%%%%%%%%%% Table I %%%%%%%%%%%%%%%%%%%
\begin{table}[h]
\begin{threeparttable}
\caption{Global Acc.(\%) and Robust Acc.(\%) of Benign Federated Learning Method}
\label{tab:2}
\setlength\tabcolsep{0pt} % make LaTeX figure out intercolumn spacing

\begin{tabular*}{\columnwidth}{@{\extracolsep{\fill}} ll cccc}
\toprule
     Global     & Robust    & Global  & Clients   & Total   & Selected    \\ 
     Acc.(\%)   & Acc.(\%)  & Rounds  & Selected  & Clients & Proportion (\%)     \\
\midrule
     {\color{green}\textbf{81.9}} & N/A & 7  & 5 & 5 & 100 \\
\midrule
     {\color{green}\textbf{82.5}} & 6.5 & 10 & 4 & 5 & 80 \\
     79.6 & 6.7 & 10 & 8 & 10 & 80 \\
     73.2 & 4.2 & 10 & 12 & 15 & 80 \\
\midrule
     {\color{green}\textbf{78.6}} & N/A & 10 & 3 & 6 & 50 \\
     76.1 & 5.0 & 10 & 4 & 8 & 50 \\
     61.5 & 3.4 & 10 & 8 & 16 & 50 \\
     2.70 & 2.7 & 10 & 10 & 20 & 50 \\
\midrule
     {\color{green}\textbf{71.8}} & 4.4 & 10 & 1 & 5 & 20 \\
     {\color{red}\textbf{48.5}} & 3.8 & 10 & 2 & 10 & 20 \\
     40.8 & 2.3 & 10 & 3 & 15 & 20 \\
     34.1 & 2.2 & 10 & 4 & 20 & 20 \\

\bottomrule
\end{tabular*}
\smallskip
\scriptsize
\RaggedRight
\end{threeparttable}
\end{table}
%%%%%%%%%%%%%%%%%%%%%%%%%%%%%%%%%%%%%%%%%%%%

\begin{figure*}[h]
    \centering
    \subfigure[]{\includegraphics[width=0.15\textwidth]{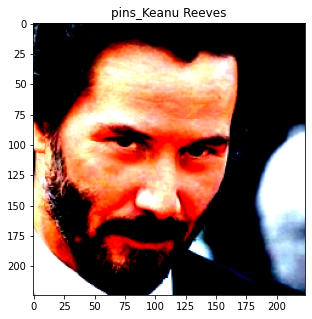}} 
    \subfigure[]{\includegraphics[width=0.15\textwidth]{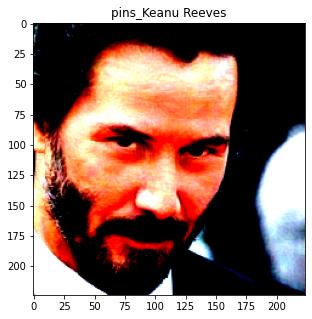}} 
    \subfigure[]{\includegraphics[width=0.15\textwidth]{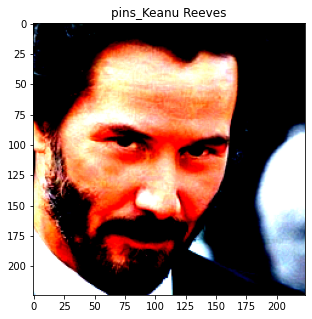}} 
    \subfigure[]{\includegraphics[width=0.15\textwidth]{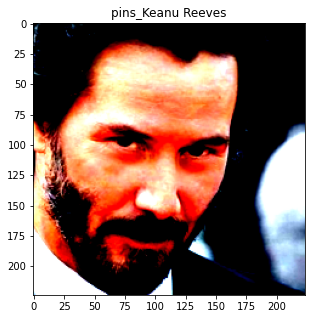}} 
    \subfigure[]{\includegraphics[width=0.15\textwidth]{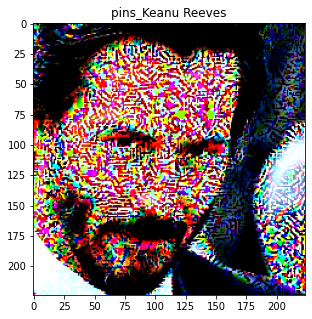}} 
    \subfigure[]{\includegraphics[width=0.15\textwidth]{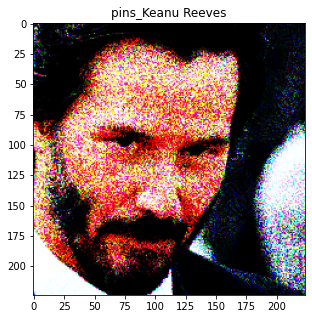}} 
    \caption{Original image \& adversarial examples generated based on the following algorithms: 
    (a) Original Image
    (b) FGSM (\(\epsilon\)=8/255)
    (c) FFGSM (\(\epsilon\)=8/255, \(\alpha\)=10/255)
    (d) Square Attack (\(\epsilon\)=8/255, \(n_{queries}\)=2000, \(n_{restarts}\)=1, loss='ce')
    (e) FGSM (\(\epsilon\)=0.9)
    (f) FFGSM (\(\epsilon\)=0.9, \(\alpha\)=10/255)
    }
    \label{fig:foobar}
\end{figure*}

For our benign federated learning experiment, we did not use any adversarial examples within our local model training process, which means that the models allocated at each individual devices will only be facing clean face images. At the same time, the test data for global accuracy measurement only consists of original images with no modifications incorporated. It is under an IID setting where the total face data is distributed equally among all clients. The number of epochs for local training was set to five for a fair experiment. Robust accuracy in this setting is the measurement of how well the global model defend against \(FGSM(\epsilon=8/255)\) created adversarial examples. Figures colored in {\color{green} \textbf{green}} are the best results obtained from each selected proportion of clients that went through federated learning. 

As shown in both Table I and Fig. 2, we can notice that less total number of clients will result in a higher global accuracy. Another trend we can see is that lowering the selected proportion of the clients for weight averaging will naturally decrease the global accuracy as well. Although the training accuracy of individual devices reached up to 99\%, the aggregated model parameter will be averaged out after each round, which results in a trend shown in Fig. 2. Another interesting finding is when proportion of client chosen is set to 20\%. The global accuracy drops significantly from 71.8\% to 48.5\% when the total clients are increased from 5 to 10. The model tends to react more sensitively in the 20\% proportion. The robust accuracy for every test cases recorded below 7\%, which means that the global was not even able to defend against majority of the FGSM-based \cite{goodfellow2014explaining} adversarial examples. 

\subsection{Robust Federated Learning (IID)}

%%%%%%%%%%%%%%%%%%% Table II %%%%%%%%%%%%%%%%%%%

\begin{table}[h]
\begin{threeparttable}
\caption{Global Acc.(\%) and Robust Acc.(\%) of Robust Federated Learning (IID). Adversarially Trained with FFGSM (\(\epsilon\)=8/255, \(\alpha\)=10/255)}

\label{tab:2}
\setlength\tabcolsep{0pt} % make LaTeX figure out intercolumn spacing

\begin{tabular*}{\columnwidth}{@{\extracolsep{\fill}} ll cccc}
\toprule
     \(n_a\)\tnote{a} & $ABR$\tnote{b}(\%)  &  Global Acc.(\%) & 
     \multicolumn{3}{c} {Robust Acc.(\%)}  \\ 
     
\cmidrule{4-6}
    &&& FGSM \cite{goodfellow2014explaining} & FFGSM \cite{wong2020fast} & Square \cite{andriushchenko2020square} \\

\midrule
     1 & 25 & 85.1 & 47.9 & 49.2 & 56.2 \\
     1 & 50 & {\color{green}\textbf{85.7}} & 54.1 & 54.2 & 60.5 \\
     1 & 75 & 85.1 & 51.6 & 52.5 & 58.6 \\

\midrule
     2 & 25 & 82.8 & 65.2 & 65.3 & 68.7 \\
     2 & 50 & {\color{green}\textbf{83.1}} & 66.7 & 67.9 & 71.0 \\
     2 & 75 & 80.7 & 67.4 & 67.9 & 68.1 \\

\midrule
     3 & 25 & {\color{green}\textbf{83.0}} & 64.9 & 65.0 & 68.45\\
     3 & 50 & 73.9 & 71.9 & 72.3 & 72.5 \\

\midrule
     4 & 25 & {\color{green}\textbf{71.5}} & 70.2 & 70.9 & 71.2 \\
     4 & 50 & {\color{red}\textbf{30.7}} & 74.1 & 75.0 & 66.6 \\

\bottomrule
\end{tabular*}

\smallskip
\scriptsize
\begin{tablenotes}
\RaggedRight
\item[a] \(n_a\): No. of clients to go through adversarial training
\item[b] $ABR$: Adversarial training batch ratio
\end{tablenotes}
\end{threeparttable}
\end{table}
%%%%%%%%%%%%%%%%%%%%%%%%%%%%%%%%%%%%%%%%%%%%

In our second experimental setting, we have investigated Method I robust federated learning system as shown in Algorithm 1. The reason why we have chosen Method I over Method II is that both average global accuracy as well as robust accuracy was recorded higher than those of Method II. Global model was trained for 10 rounds with 5 number of epochs for local devices and clients chosen for robust training were trained with FFGSM (\(\epsilon\)=8/255, \(\alpha\)=10/255) created adversarial examples. We also set the total number of clients to be 5 for an efficient experiment procedure. During evaluation, the global model was also evaluated based on the following adversarial attack algorithms: 
\begin{itemize}
    \item FGSM (\(\epsilon\)=8/255)
    \item FFGSM (\(\epsilon\)=8/255, \(\alpha\)=10/255)
    \item Square Attack (\(\epsilon\)=8/255, \(n_{queries}\)=2000, \(n_{restarts}\)=1, loss='ce')
\end{itemize}

We have also added a black-box model setting to give a more realistic situation by creating the adversarial examples based on a third-party model. In this case, we have used the gradient information from pre-trained AlexNet \cite{krizhevsky2017imagenet}. The model consists of an entirely different parameter compared to that of ResNet-34 \cite{he2016deep}. Fig. 3 illustrates the adversarial examples generated based on AlexNet \cite{krizhevsky2017imagenet}. As we may notice, there is a imperceptible contrast between the original image and the examples created with only a slight perturbation ((b), (c), (d) of Fig. 3). We can detect the noises only when the \(\epsilon\) value is set relatively higher ((e) and (f) of Fig. 3). 

Moreover, new feature \(ABR(\%\)) is included in our experiment as well. As mentioned earlier, when a specific client ID is chosen to process adversarial training in the given round, \(ABR\) decides the proportion of adversarial examples to be used in a local training batch. Finding the balance between \(n_a\) and \(ABR\) to preserve high accuracies against both clean and perturbed data is one of the key features of the following experiments. 

Global accuracy generally tend to decrease while robust accuracy showed an increasing trend when both \(n_a\) and \(ABR\) increased. The accuracies reach to a certain point with an average of 71\% after increasing \(n_a\) up to either 3 or 4. But, the global accuracy dropped remarkably from 71.5\% to 30.7\% when \(ABR\) was raised from 25 to 50 in a \(n_a=4\) scenario, while most robust accuracies were either maintained or escalated to a certain level. According to the results, the perfect balance will be the case when \(n_a=3\) and \(ABR=50\), where global accuracy reached up to 73.9\% while the robust accuracy of the global model against all the other three attack algorithms converged to a 70\% range. Such result further proves the transferability of the adversarial examples and the robustness of DNNs. 

\subsection{Pixel Modification (Non-IID)}

\begin{figure}
    \centering
    \subfigure[]{\includegraphics[width=0.2\textwidth]{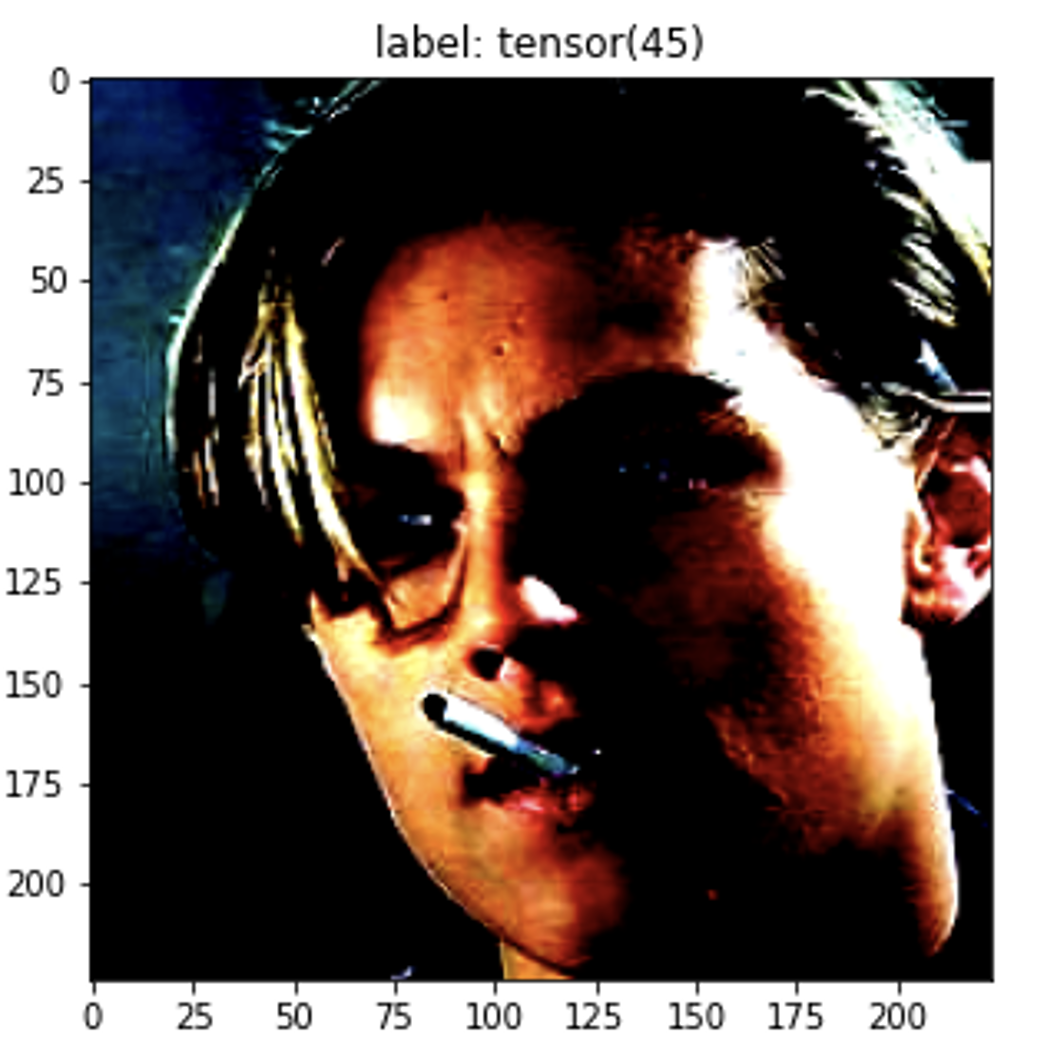}} 
    \subfigure[]{\includegraphics[width=0.2\textwidth]{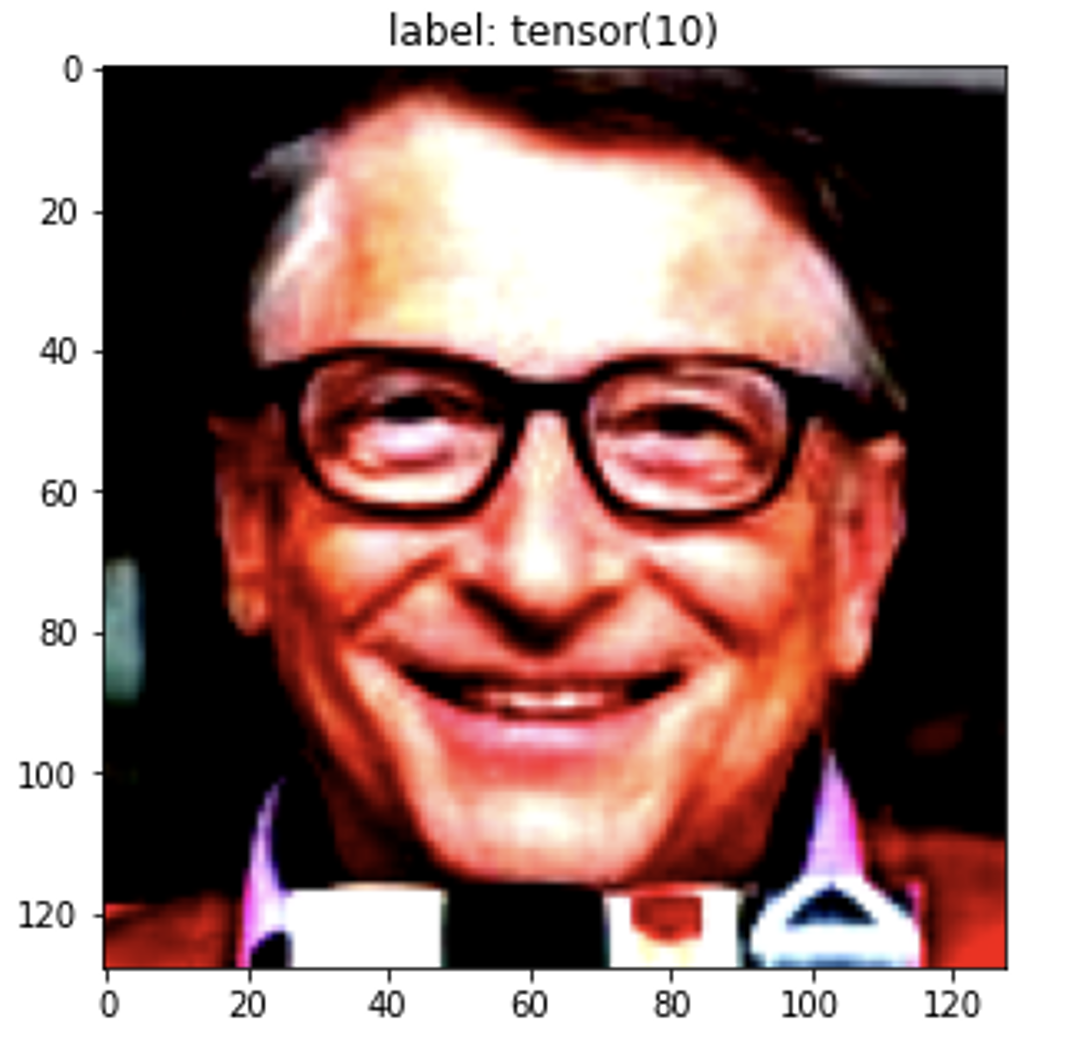}} 
    \caption{(a) Original Image (\(3\times224\times224\)) (b) Pixel Modified Image (\(3\times128\times128\))}
    \label{fig:foobar}
\end{figure}

In this section, we investigated the robustness of the model in a Non-IID federated learning environment where both the size and the distribution of the data in individual devices differ from each other. As for our default setting for the remaining experiments, we set a total of 5 clients and randomly select 4 devices amongst the 5. The client IDs for data manipulation will be chosen in the very beginning of our experiment, which implies that those devices will only contain the modified face images. In this case, we degraded the pixels from \(3 \times 224 \times 224\) to \(3 \times 128 \times 128\) in order to observe how the global model reacts. General overview of FLATS in Non-IID setting is illustrated in Fig. 5. 

\begin{figure}[htp]
    \centering
    \includegraphics[width=260pt]{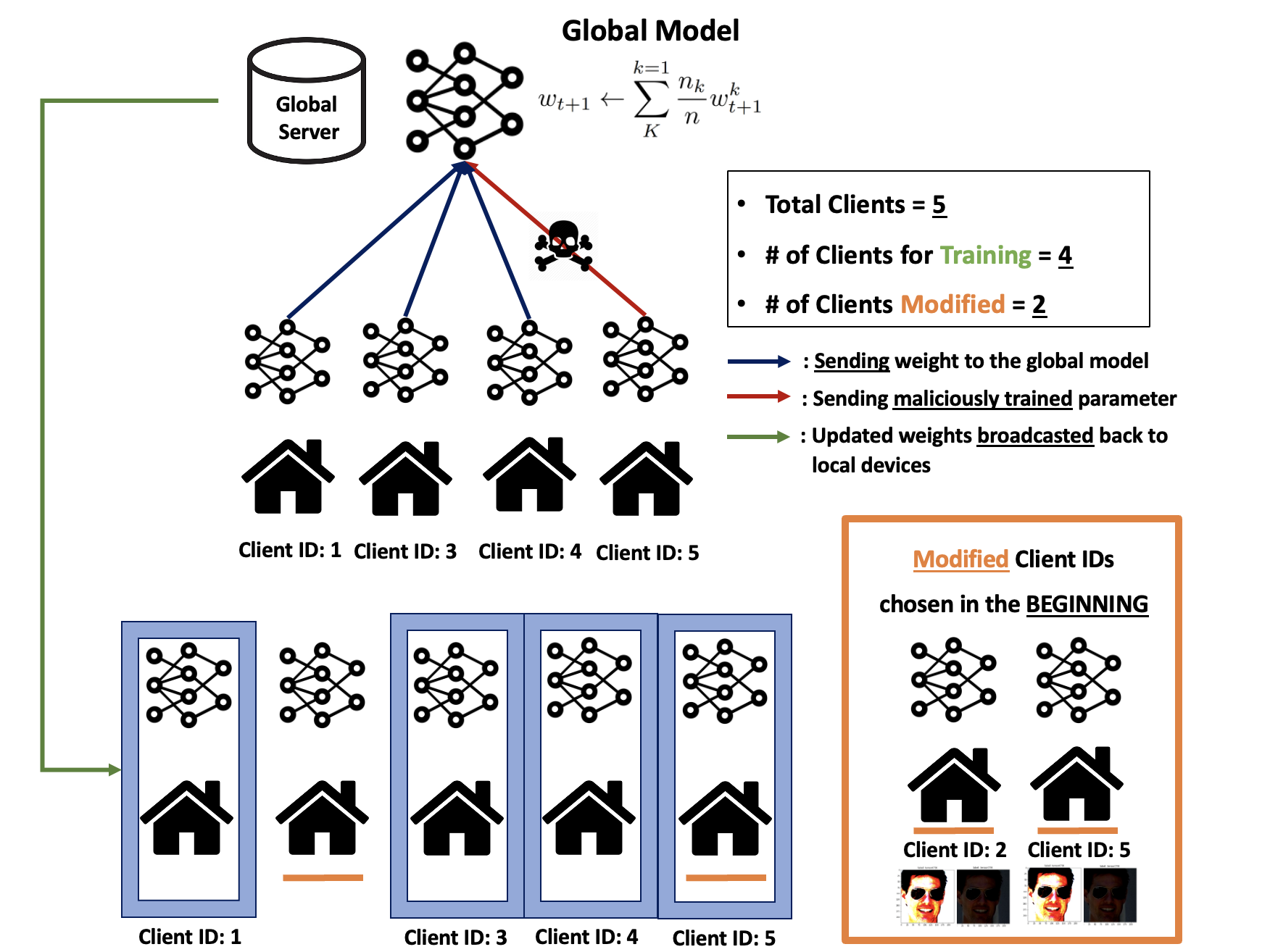}
    \caption{Overview of FLATS in Non-IID Setting}
    \label{fig:Robust_FL_2}
\end{figure}

%%%%%%%%%%%%%%%%% Table III %%%%%%%%%%%%%%%%%

\begin{table}[h]
\begin{threeparttable}
\caption{Global Acc.(\%) and Robust Acc.(\%) of Robust Federated Learning (Non-IID). Two random clients Pixel Modified to \(3 \times 128 \times 128\)}

\label{tab:2}
\setlength\tabcolsep{0pt} % make LaTeX figure out intercolumn spacing

\begin{tabular*}{\columnwidth}{@{\extracolsep{\fill}} ll cccc}
\toprule
     \(n_a\)\tnote{a} & $ABR$\tnote{b}(\%)  &  Global Acc.(\%) & 
     \multicolumn{3}{c} {Robust Acc.(\%)}  \\ 
     
\cmidrule{4-6}
    &&& FGSM \cite{goodfellow2014explaining} & FFGSM \cite{wong2020fast} & Square \cite{andriushchenko2020square} \\

\midrule
     1 & 50 & {\color{red}\textbf{77.3}} & {\color{red}\textbf{42.2}} & {\color{red}\textbf{41.7}} & {\color{red}\textbf{48.4}} \\
     2 & 50 & 76.5 & 60.7 & 61.6 & 64.5 \\
     3 & 50 & 61.6 & 63.7 & 64.5 & 64.1 \\
     4 & \textbf{25} & 57.6 & 61.7 & 63.2 & 62.4 \\

\bottomrule
\end{tabular*}

\smallskip
\scriptsize
\begin{tablenotes}
\RaggedRight
\item[a] \(n_a\): No. of clients to go through adversarial training
\item[b] $ABR$: Adversarial training batch ratio
\end{tablenotes}
\end{threeparttable}
\end{table}

%%%%%%%%%%%%%%%%%%%%%%%%%%%%%%%%%%%%%%%%%%%%%%%%%%%%%%%%%%%%%%%%%%%

\begin{figure}
    \centering
    \subfigure[]{\includegraphics[width=0.24\textwidth]{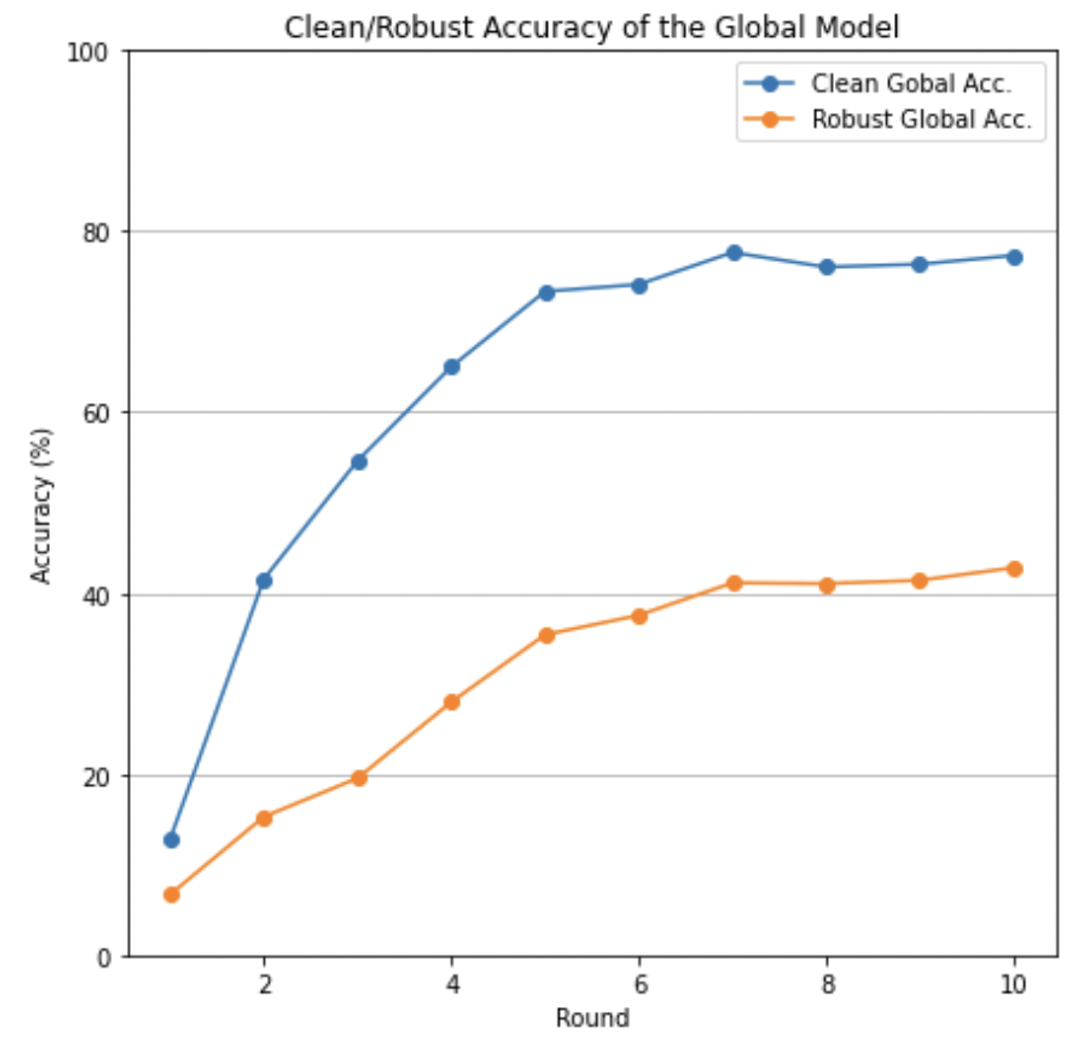}} 
    \subfigure[]{\includegraphics[width=0.24\textwidth]{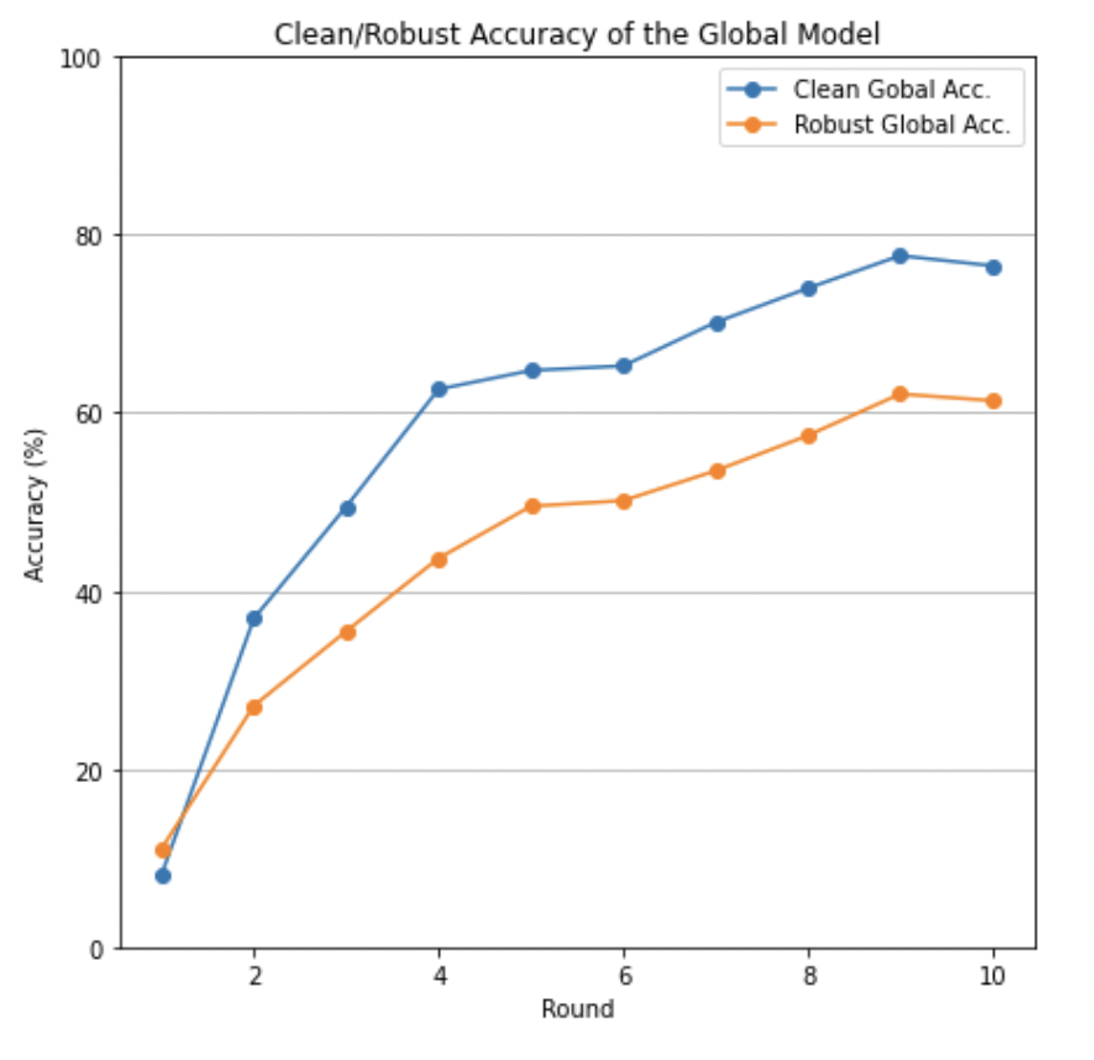}} 
    \subfigure[]{\includegraphics[width=0.24\textwidth]{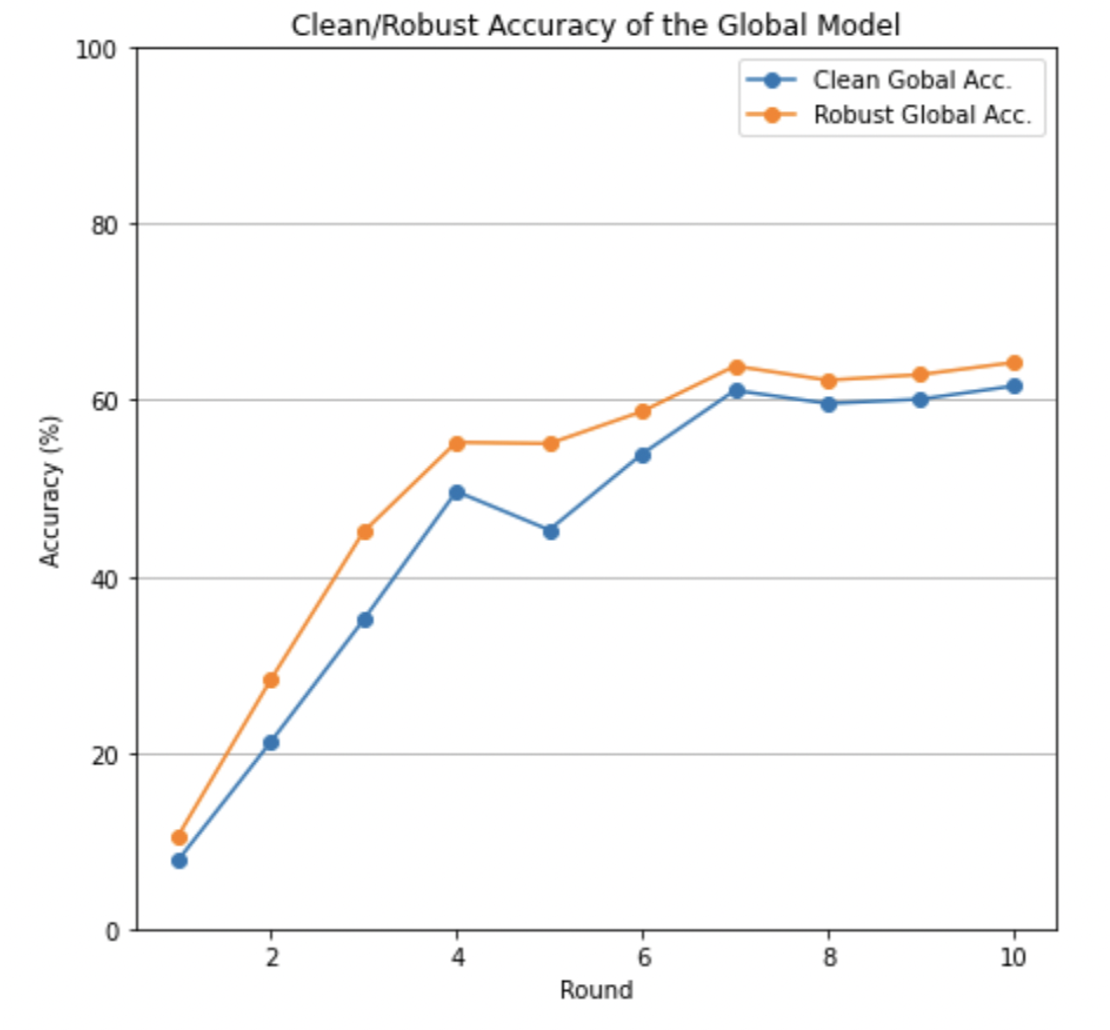}} 
    \subfigure[]{\includegraphics[width=0.24\textwidth]{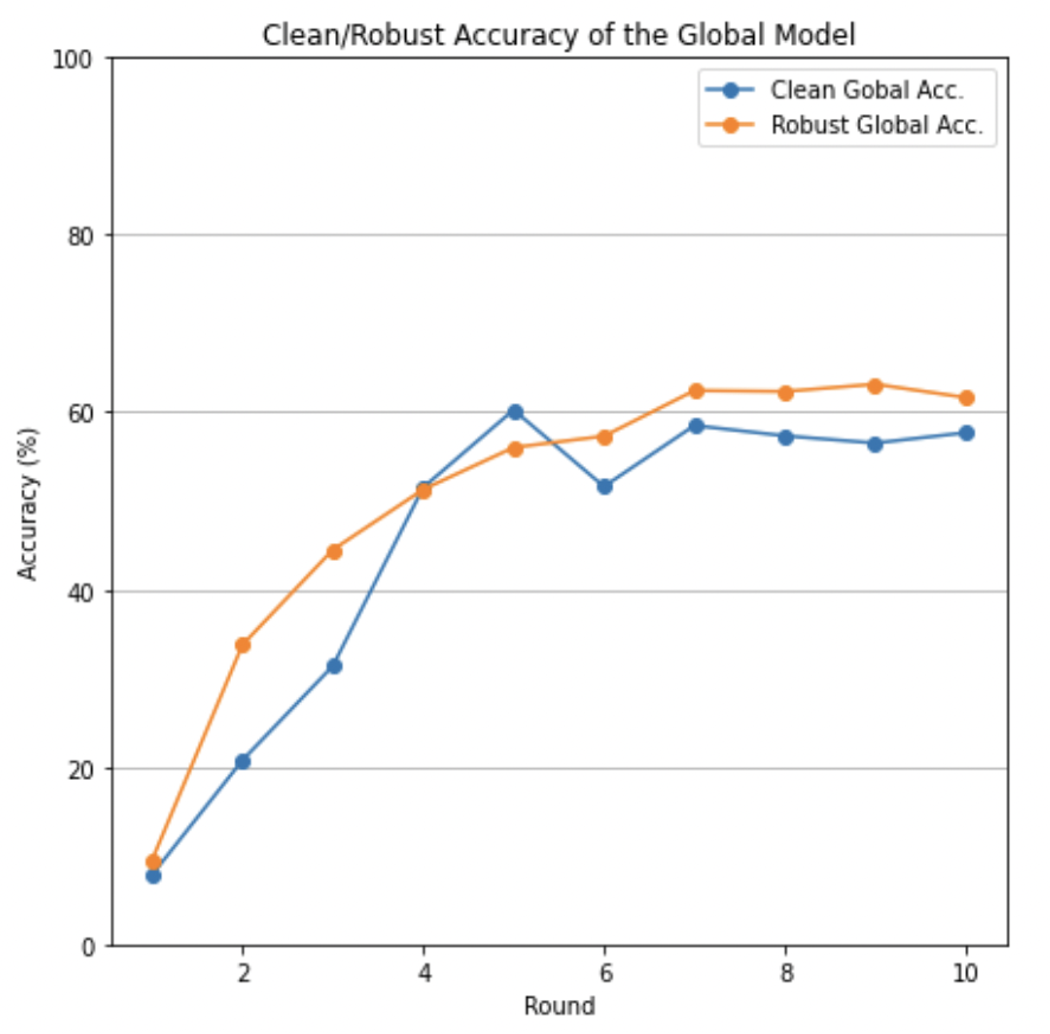}} 
    \caption{Clean \& Robust Acc.(\%) of Pixel Modified Robust Federated Learning in each round. Robust Acc.(\%) against FFGSM (\(\epsilon\)=8/255, \(\alpha\)=10/255) (a) \(n_a\)=1 (b) \(n_a\)=2 (c) \(n_a\)=3 (d) \(n_a\)=4}
    \label{fig:foobar}
\end{figure}

%%%%%%%%%%%%%%%%%%%%%%%%%%%%%%%%%%%%%%%%%%%%%%%%%%%%%%%%%%%%%%%%%%%

From Fig. 4, we may notice that there is only a subtle distinction between the original face image and the pixel modified image. However, the global model reacts quite responsively as we can observe from Table III. \(ABR\) was chosen based on the highest global accuracy recorded in each different \(n_a\), which are highlighted in {\color{green} \textbf{green}} in Table II.

In general, both global accuracy and robust accuracy dropped to a certain degree. However, they all decreased more notably when \(n_a\)=1 compared to other settings. Furthermore, if we refer to the graphs illustrated in Fig. 5, the first observation we would like to emphasize is that the degree of increase of the robust accuracy trend is remarkably higher to that of decrease in the global accuracy. From this result, we may argue that the model tend to become robust even against adversarial attacks once pixel modified images are included in the whole training process. Secondly, from (c) and (d) of Fig. 6, global accuracy drops approximately between rounds 4 to 6. However, it recuperates its classification ability immediately in the following rounds and maintains its increasing trend. It demonstrates that even DNNs can overcome or recover from the mistakes they have been made, or repair their malfunctioning neurons, just like the muscle memory in our body. 
 
\subsection{Covering "Eye" Area (Non-IID)}
%%%%%%%%%%% Table IV %%%%%%%%%%%%%%%%

\begin{table}[h]
\begin{threeparttable}
\caption{Global Acc.(\%) and Robust Acc.(\%) of Robust Federated Learning (Non-IID). Two random clients consist data with "eye area" covered}

\label{tab:2}
\setlength\tabcolsep{0pt} % make LaTeX figure out intercolumn spacing

\begin{tabular*}{\columnwidth}{@{\extracolsep{\fill}} ll cccc}
\toprule
     \(n_a\)\tnote{a} & $ABR$\tnote{b}(\%)  &  Global Acc.(\%) & 
     \multicolumn{3}{c} {Robust Acc.(\%)}  \\ 
     
\cmidrule{4-6}
    &&& FGSM \cite{goodfellow2014explaining} & FFGSM \cite{wong2020fast} & Square \cite{andriushchenko2020square} \\

\midrule
     1 & 50 & 73.6 & 39.3 & 40.6 & 47.3 \\
     2 & 50 & 77.4 & 60.6 & 61.6 & 64.9 \\
     3 & 50 & 69.1 & 60.6 & 61.3 & 63.0 \\
     4 & \textbf{25} & 63.8 & 63.1 & 63.5 & 63.4 \\

\bottomrule
\end{tabular*}

\smallskip
\scriptsize
\begin{tablenotes}
\RaggedRight
\item[a] \(n_a\): No. of clients to go through adversarial training
\item[b] $ABR$: Adversarial training batch ratio
\end{tablenotes}
\end{threeparttable}
\end{table}
%%%%%%%%%%%%%%%%%%%%%%%%%%%%%%%%%%%%%%%%%%%%%%%%%%%%%%%%%%%%%%%%%%%%%%%%%%%

%%%%%%%%%%%%%%%%%%%%%%%%%%%%%%%%%%%%%%%%%%%%%%%%%%%%%%%%%%%%%%%%%%%%%%%%%%%
\begin{figure}
    \centering
    \subfigure[]{\includegraphics[width=0.2\textwidth]{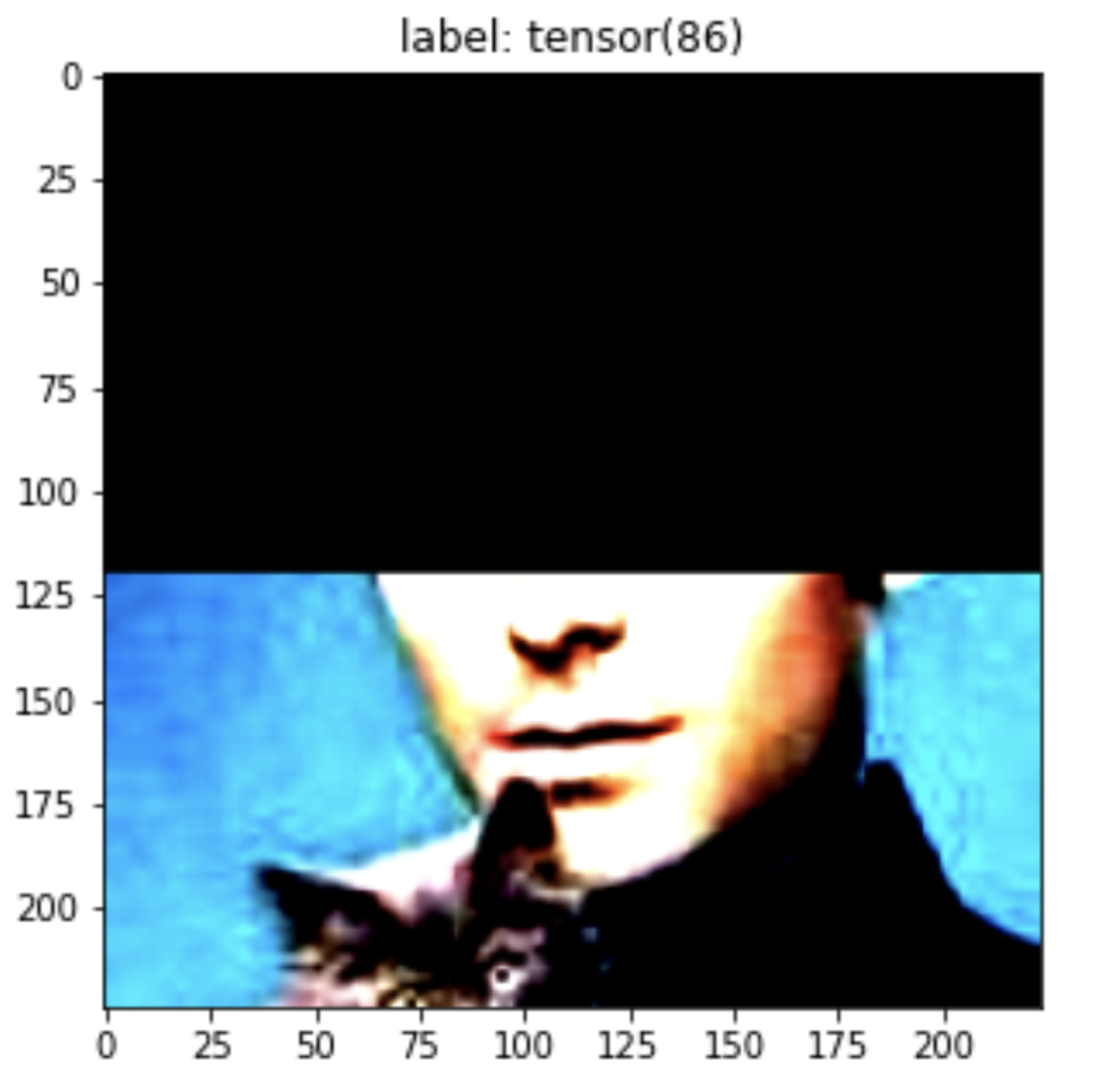}} 
    \subfigure[]{\includegraphics[width=0.2\textwidth]{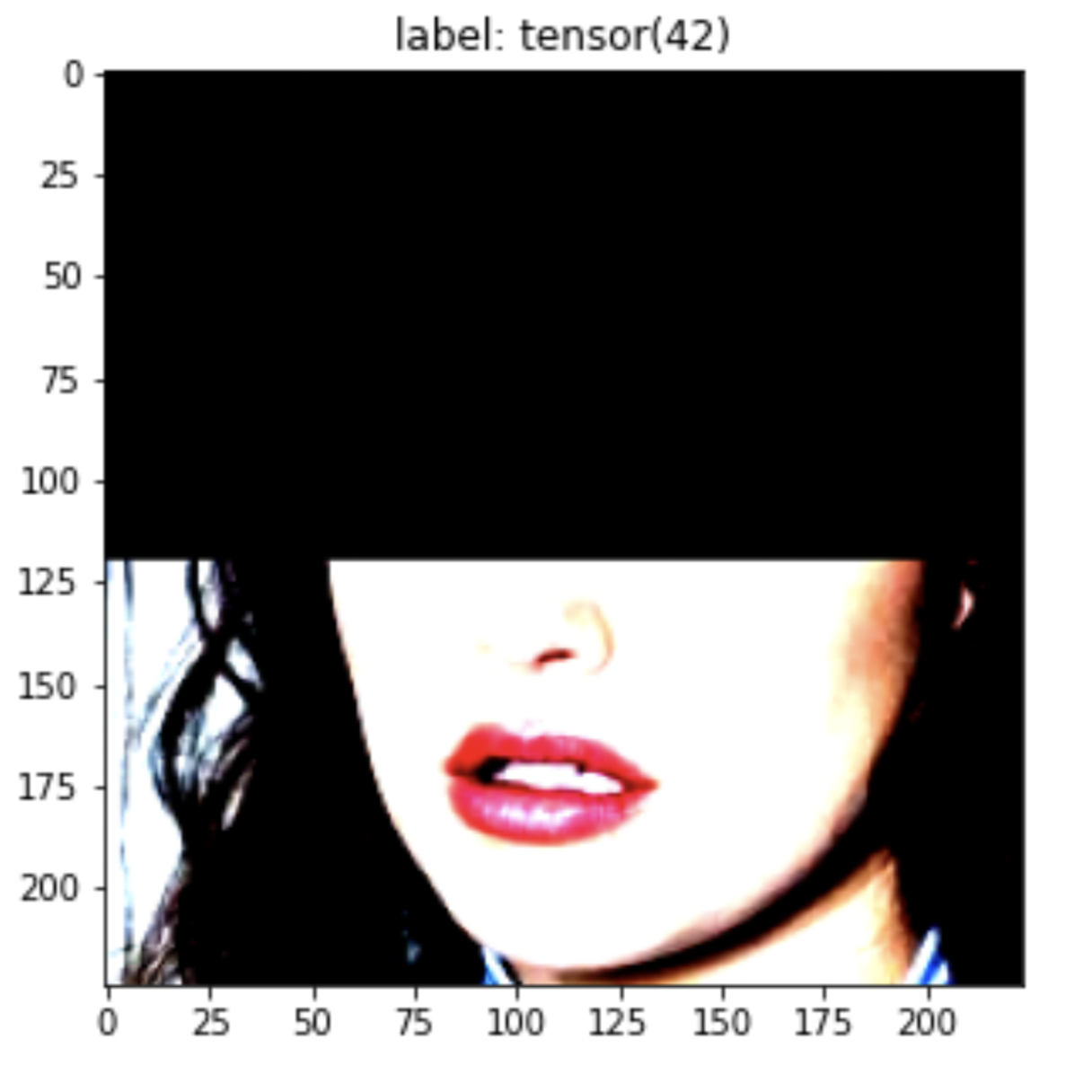}} 
    \caption{(a) "Eye Area" Covered Face Image 1 (b) "Eye Area" Covered Face Image 2}
    \label{fig:foobar}
\end{figure}
%%%%%%%%%%%%%%%%%%%%%%%%%%%%%%%%%%%%%%%%%%%%%%%%%%%%%%%%%%%%%%%%%%%%%%%%%%%

Our eyes are considered to be one of the unique parts of our faces for personal identification. We wanted to observe how such obstacles impact the classification result of our global face recognition model. The method can also be considered as a backdoor attack, which have been investigated in \cite{goswami2018unravelling, wenger2021backdoor, chen2017targeted} in a similar manner. Moreoever, as for protection for the smart home security, any intruders might cover their eyes in order to break through the automatic face recognition system. Such investigated result will demonstrate how the model deals with such disturbance. To ensure that the eye area will be covered in every images that are chosen to be modified, we have erased the entire pixels from image height of 0 to 120 as illustrated in Fig. 7. As a result, there was only a mere difference in both accuracies compared to the those in the pixel modified setting.  

\subsection{Brightness Modification (Non-IID)}

%%%%%%%%%%%%%%%%%%%%%% Table V %%%%%%%%%%%%%%%%%%%%%%%%%%%%%%%%%%%
\begin{table}[h]
\begin{threeparttable}
\caption{Global Acc.(\%) and Robust Acc.(\%) of Robust Federated Learning (Non-IID). Two random clients with brightness modified}

\label{tab:2}
\setlength\tabcolsep{0pt} % make LaTeX figure out intercolumn spacing

\begin{tabular*}{\columnwidth}{@{\extracolsep{\fill}} ll ccccc}
\toprule
     \(n_a\)\tnote{a} & $ABR$\tnote{b}(\%) & $BR$\tnote{c} & $GA$\tnote{d}(\%) &
     \multicolumn{3}{c} {Robust Acc.(\%)}  \\ 
     
\cmidrule{5-7}
    &&&& FGSM \cite{goodfellow2014explaining} & FFGSM \cite{wong2020fast} & Square \cite{andriushchenko2020square} \\

\midrule
     1 & 50 & 0.15 & 79.9 & {\color{green}\textbf{57.9}} & {\color{green}\textbf{59.1}} & {\color{green}\textbf{63.3}} \\
     2 & 50 & 0.15 & 65.6 & {\color{green}\textbf{70.6}} & {\color{green}\textbf{71.3}} & 70.3 \\
     3 & 50 & 0.15 & {\color{red}\textbf{23.6}} & {\color{green}\textbf{74.1}} & {\color{green}\textbf{75.2}} & 65.4 \\
     4 & 25 & 0.15 & 67.5 & {\color{green}\textbf{72.5}} & {\color{green}\textbf{73.5}} & {\color{green}\textbf{72.5}} \\

\midrule
     1 & 50 & 2.30 & 72.6 & 60.4 & 61.3 & 63.8 \\
     2 & 50 & 2.30 & 57.4 & 67.8 & 68.4 & 66.4 \\
     3 & 50 & 2.30 & {\color{red}\textbf{18.7}} & 68.7 & 69.6 & 60.1 \\
     4 & 50 & 2.30 & {\color{red}\textbf{10.4}} & 69.0 & 69.5 & 58.3 \\
     4 & 25 & 2.30 & 39.4 & 69.4 & 70.0 & 64.3 \\

\bottomrule
\end{tabular*}

\smallskip
\scriptsize
\begin{tablenotes}
\RaggedRight
\item[a] \(n_a\): No. of clients to go through adversarial training
\item[b] $ABR$: Adversarial training batch ratio
\item[c] $BR$: Brightness Ratio (0.15: dark / 2.30: bright)
\item[d] $GA$: Global Accuracy
\end{tablenotes}
\end{threeparttable}
\end{table}
%%%%%%%%%%%%%%%%%%%%%%%%%%%%%%%%%%%%%%%%%%%%%%%%%%%%%%%%%%%%%%%%%%%

\begin{figure*} [h]
    \centering
    \subfigure[]{\includegraphics[width=0.2\textwidth]{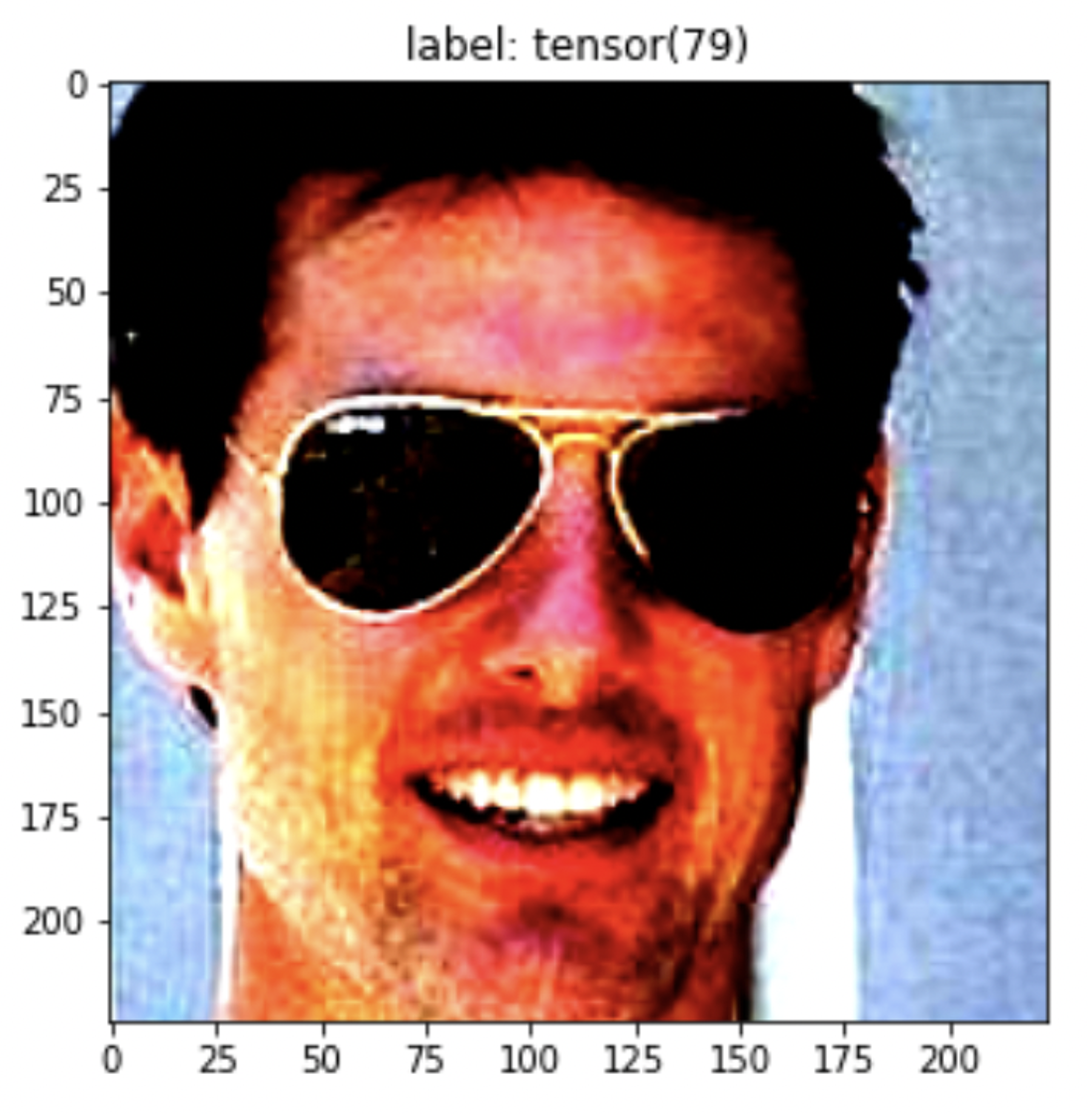}} 
    \subfigure[]{\includegraphics[width=0.2\textwidth]{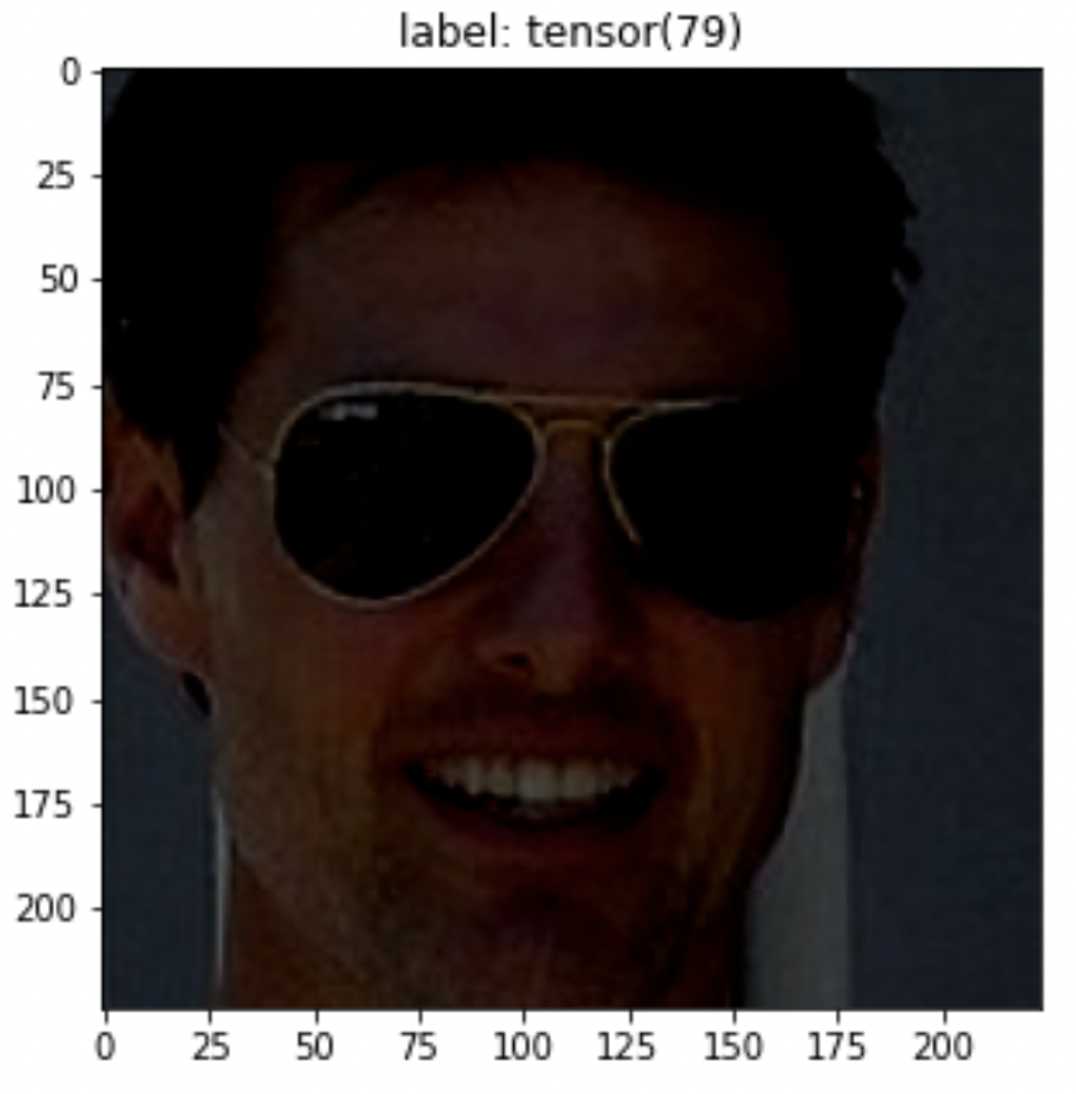}} 
    \subfigure[]{\includegraphics[width=0.2\textwidth]{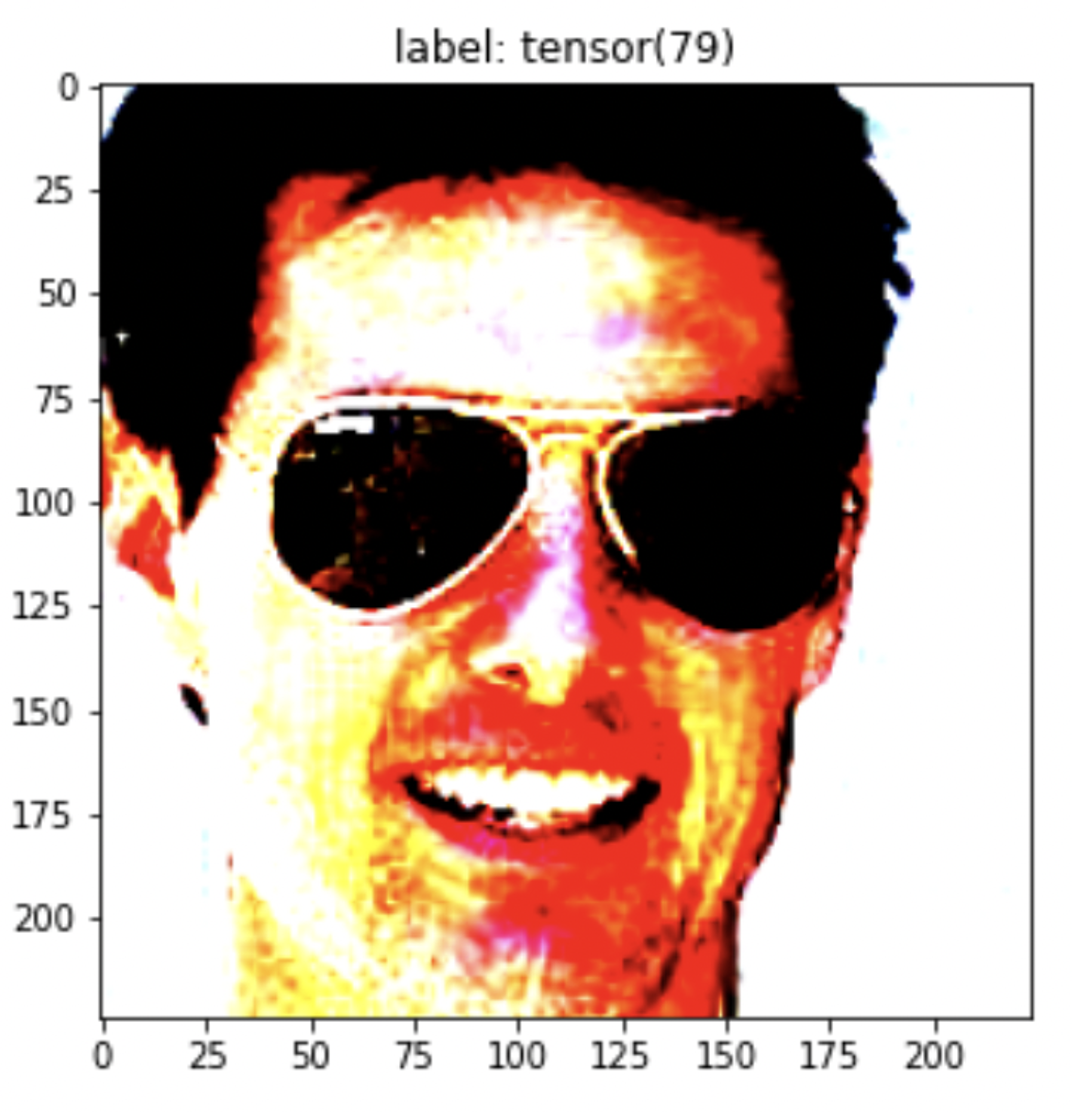}} 
    \caption{(a) Original Image (b) \(BR\): 0.15 (c) \(BR\): 2.30}
    \label{fig:foobar}
\end{figure*}

\begin{figure}
    \centering
    \subfigure[]{\includegraphics[width=0.24\textwidth]{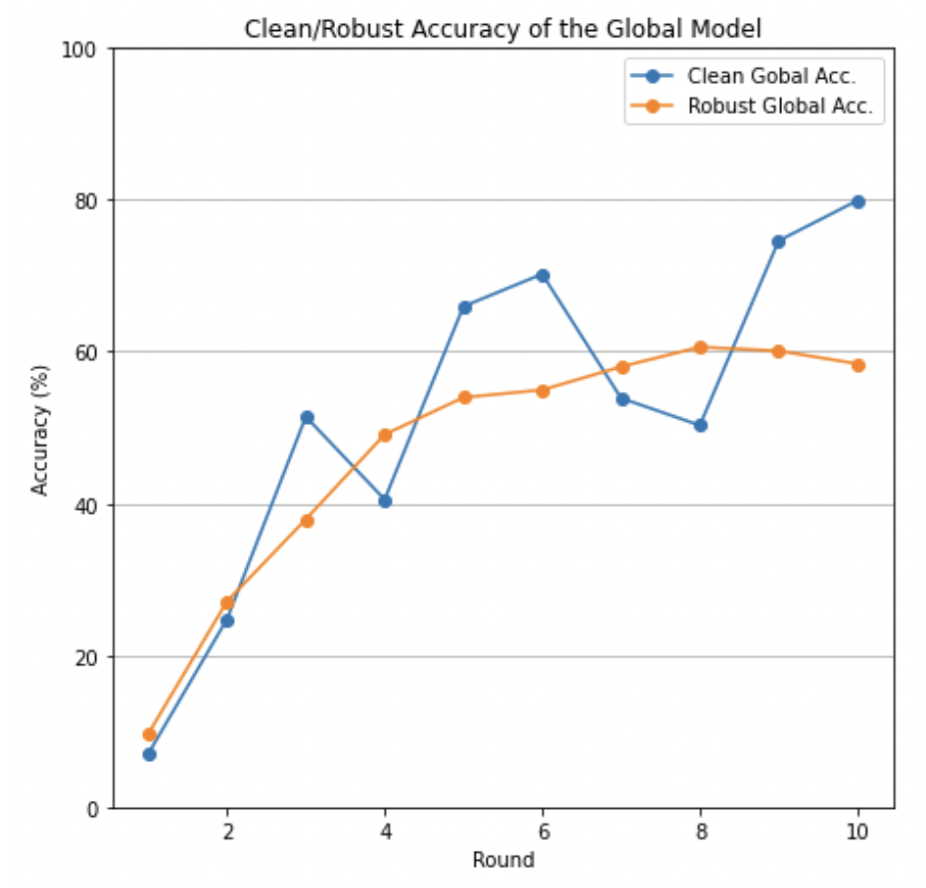}} 
    \subfigure[]{\includegraphics[width=0.24\textwidth]{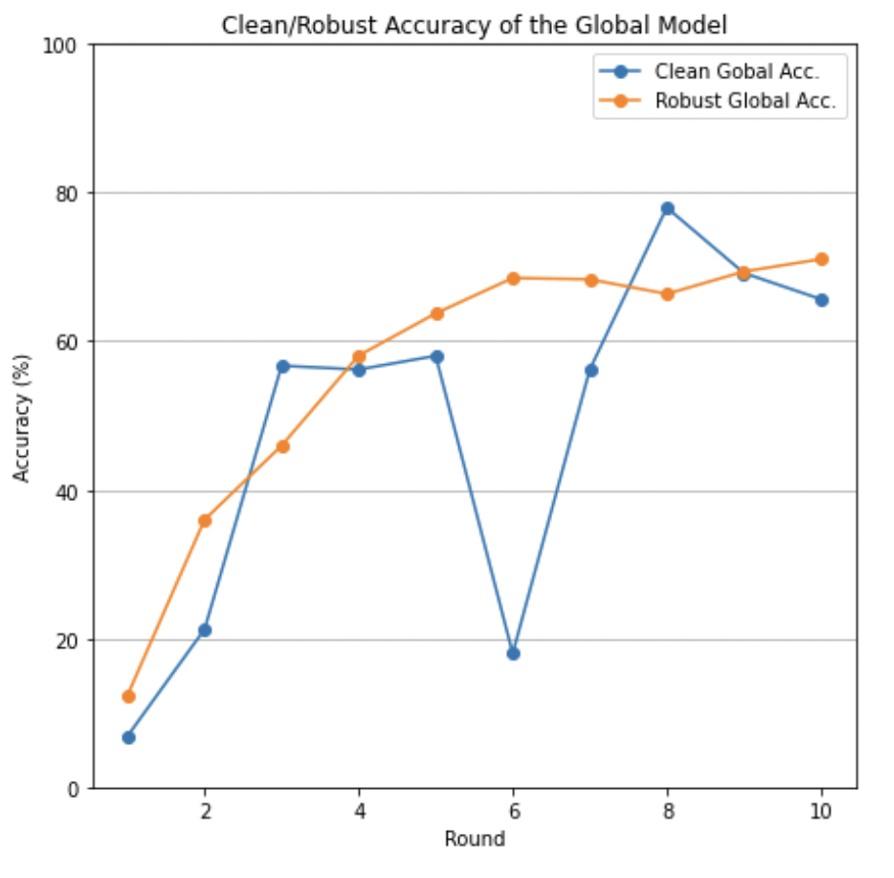}} 
    \subfigure[]{\includegraphics[width=0.24\textwidth]{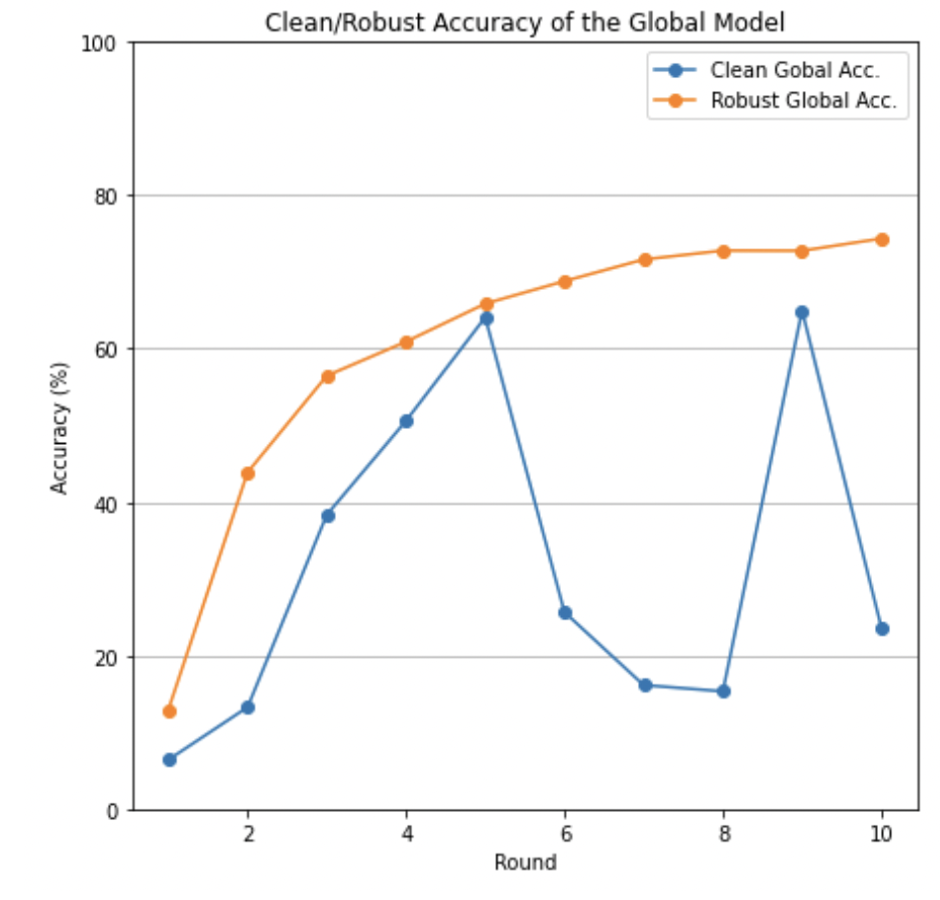}} 
    \subfigure[]{\includegraphics[width=0.24\textwidth]{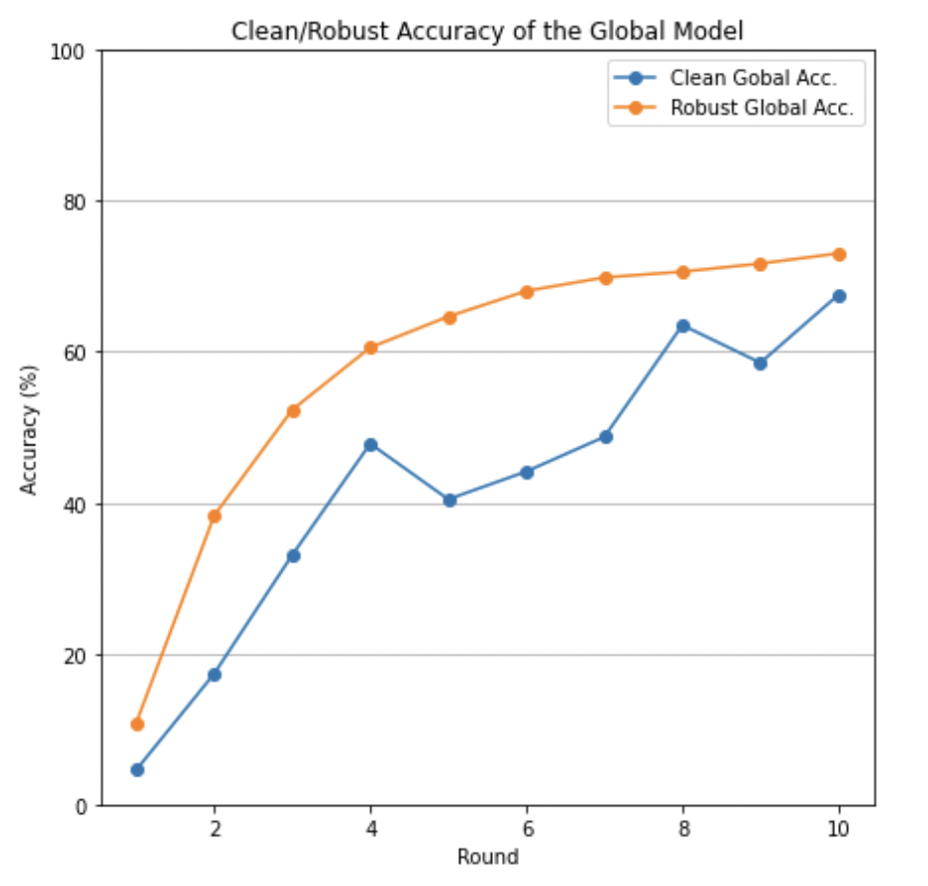}} 
    \caption{Clean \& Robust Acc.(\%) of Robust Federated Learning (Two Random Clients with "Dark" Images) (a) \(n_a\)=1 (b) \(n_a\)=2 (c) \(n_a\)=3 (d) \(n_a\)=4}
    \label{fig:foobar}
\end{figure}

\begin{figure}
    \centering
    \subfigure[]{\includegraphics[width=0.24\textwidth]{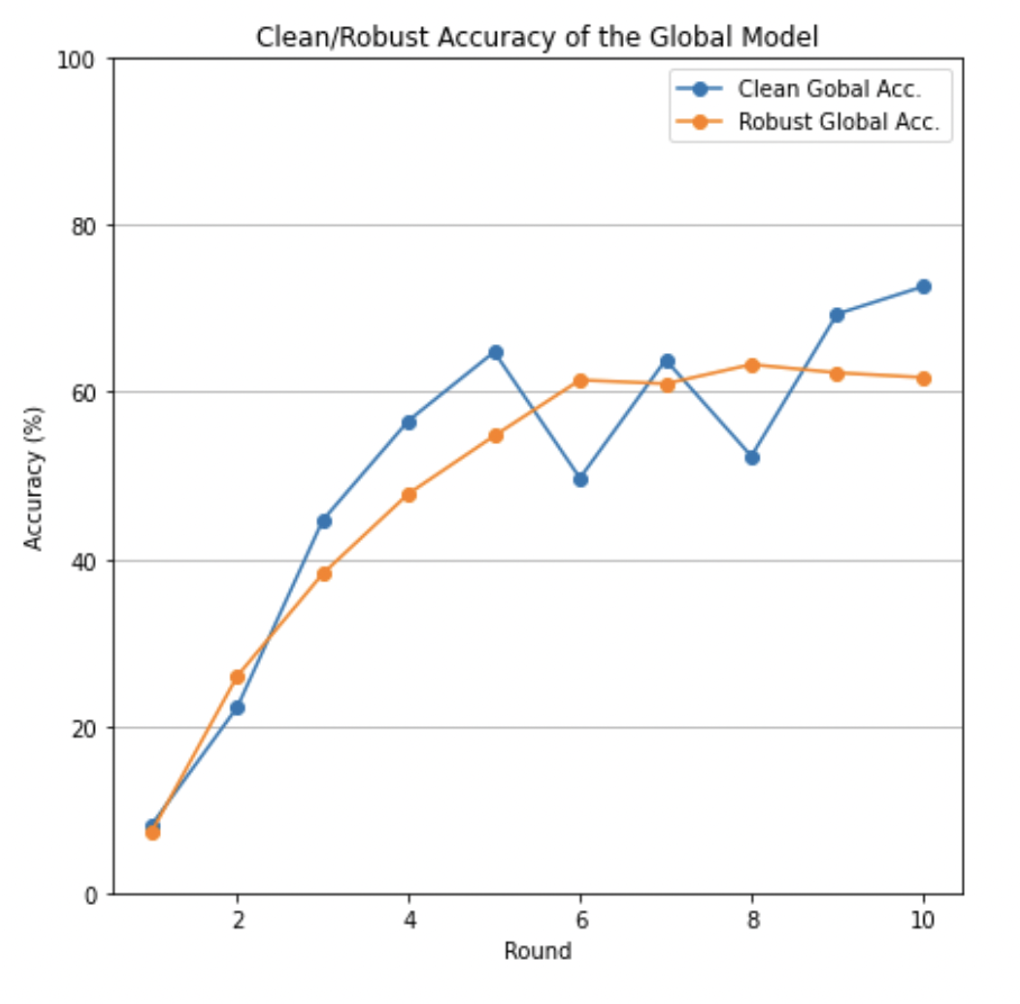}} 
    \subfigure[]{\includegraphics[width=0.24\textwidth]{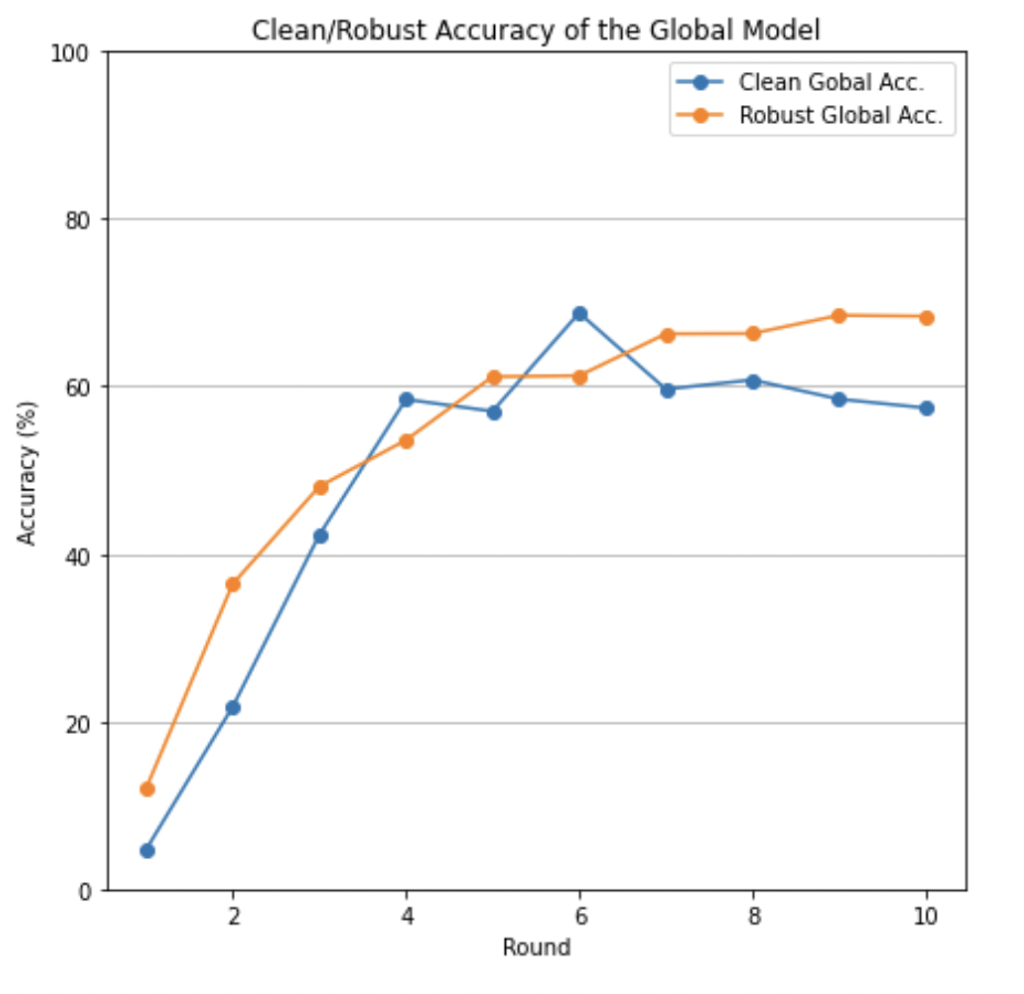}} 
    \subfigure[]{\includegraphics[width=0.24\textwidth]{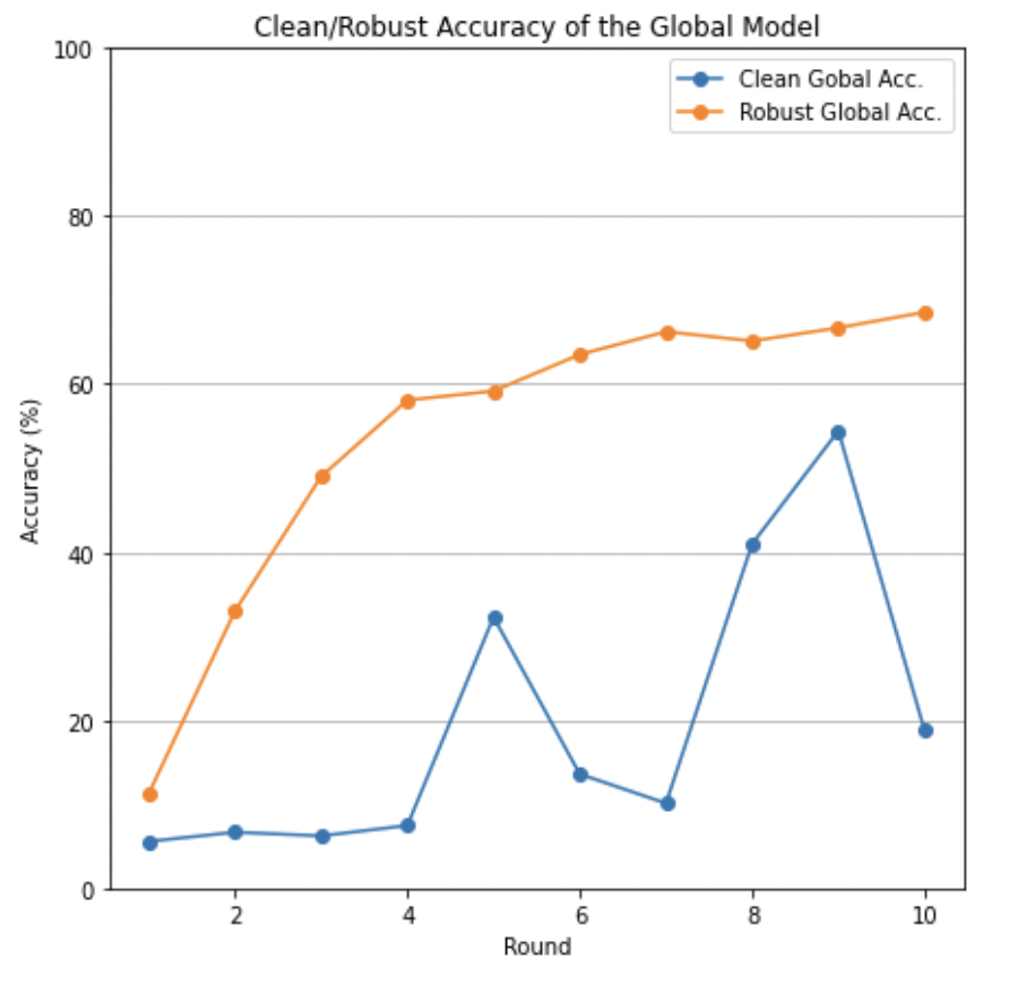}} 
    \subfigure[]{\includegraphics[width=0.24\textwidth]{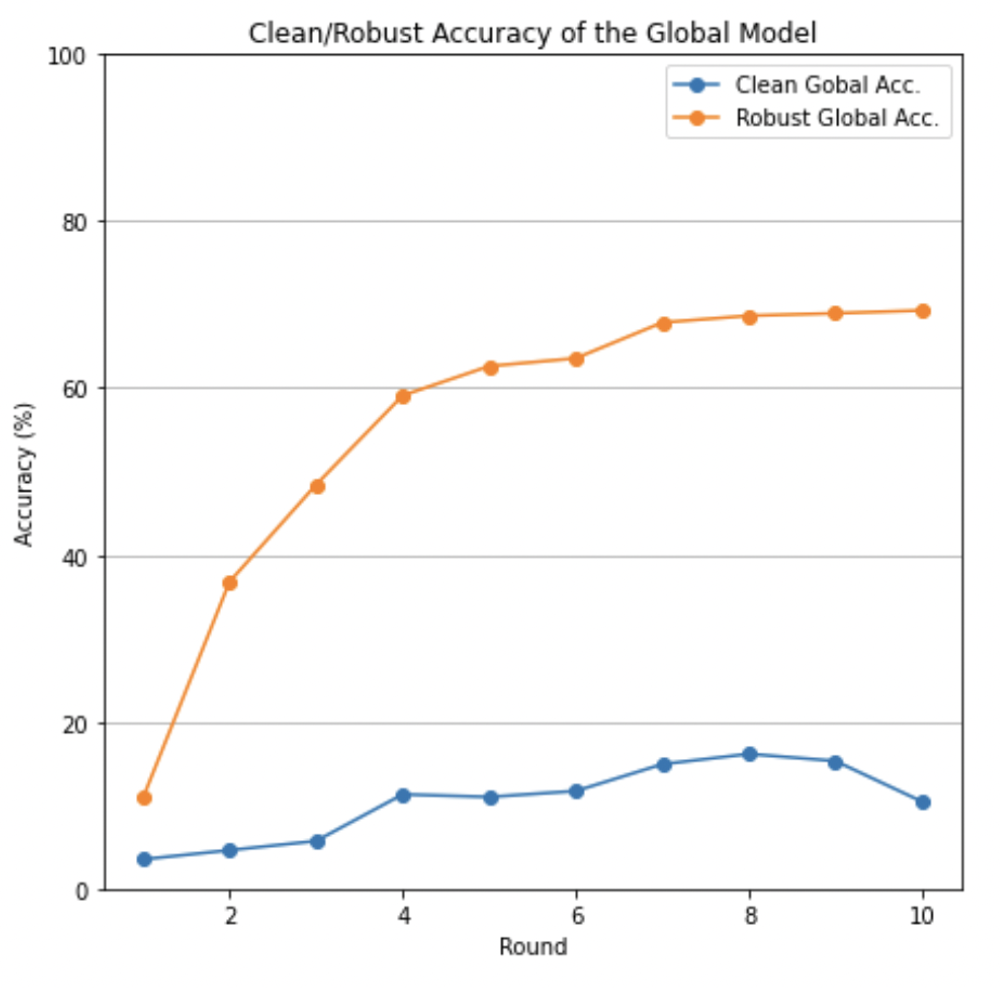}} 
    \caption{Clean \& Robust Acc.(\%) of Robust Federated Learning (Two Random Clients with "Bright" Images) (a) \(n_a\)=1 (b) \(n_a\)=2 (c) \(n_a\)=3 (d) \(n_a\)=4}
    \label{fig:foobar}
\end{figure}
%%%%%%%%%%%%%%%%%%%%%%%%%%%%%%%%%%%%%%%%%%%%%%%%%%%%%%%%%%%%%%

We have included brightness modified face images in two randomly selected clients every round during our robust federated learning training process. Fig. 8 shows the corresponding images depending on the Brightness Ratio (\(BR\)). \(BR\) of the original image is set to 1 in default. Numbers highlighted in {\color{green} \textbf{green}} in Table V represents the accuracies that are improved from the original robust accuracy that are listed in Table II, where no modifications were added. On the other hand, figures highlighted in {\color{red} \textbf{red}} are the global accuracies that dropped strikingly compared to those from IID Robust Federated Learning. 

When the "dark" images were included in the training process, chaotic fluctuations were exhibited regardless of the value of \(n_a\). In fact, the increase in global accuracy became more stable when \(n_a\) = 4. On the other hand, when we added "bright" images in two randomly selected clients, both global and robust accuracy showed a stable trend when \(n_a\) = 1 and \(n_a\) = 2. Although a minor fluctuation was presented in (a) and (b) in Fig. 9, both global models were able to recover its classification ability just like the graphs showed in Fig. 6. However, the global standard accuracy either oscillated or remained below 20\% when \(n_a\) = 3 and \(n_a\) = 4. 

The most fascinating finding in this specific setting, whether the images were adjusted brighter or darker, is the consistent trend of the robust accuracy and the fluctuating global accuracy. If the brightness modified images are embedded within the federated learning training process, we may imply that the neural network becomes confused in the evaluation phase when it meets the original clean images. However, the model seems to be familiar with the adversarial examples, which led to a fairly high robust accuracy in any given environment.The aggregated model preserves its robustness with an increasing trend in classification accuracy even with a starving federated data distributed to each devices. 

As for further explanation of such result, both adversarial examples and brightness modified images share an interchangeable attribute where their individual pixels are added with a specific value (whether it is an integer or a perturbation rate), which means that the global model was able to face more often with these perturbed images not even during the adversarial training process, but also during standard training. If we are able to find the optimal intersecting point of hyperparameters where we can sustain both high global and robust accuracy of our aggregated model, we may firmly argue that adversarial training should be a mandatory procedure, which needs to be included in the standard training of any neural network. 

%%%%%%%%%%%%%%%%%%%%%%%%%%%%%%%%%%%%%%%%%%%%%%%%%%%%%%%%%%%%%%
\subsection{Augmented Test Data (Non-IID)}
%%%%%%%%%%%%%%%%%%% Table VI %%%%%%%%%%%%%%%%%%%%%%%%%%%%%%%
\begin{table}[h]
\begin{threeparttable}
\caption{Robust Federated Learning (Non-IID) with two random clients consist of "dark" images. Evaluated on augmented test data}

\label{tab:2}
\setlength\tabcolsep{0pt} % make LaTeX figure out intercolumn spacing

\begin{tabular*}{\columnwidth}{@{\extracolsep{\fill}} ll ccccc}
\toprule
     \(n_a\)\tnote{a} & $ABR$\tnote{b}(\%) & $TDT$\tnote{c} & $GA$\tnote{d}(\%) &
     \multicolumn{3}{c} {Robust Acc.(\%)}  \\ 
     
\cmidrule{5-7}
    &&&& FGSM & FFGSM & Square\\

\midrule
     1 & 50 & Bright + Clean        & 54.4 & 57.0 & 56.9 & 61.5 \\
     1 & 50 & Bright + Dark + Clean & 55.9 & 64.5 & 66.5 & 67.7 \\
     1 & 50 & Dark + Clean          & 61.2 & 56.7 & 58.2 & 61.9 \\
\midrule
     2 & 50 & Bright + Clean        & 36.5 & 71.7 & 72.5 & 63.8 \\
     2 & 50 & Bright + Dark + Clean & 57.8 & 65.2 & 66.1 & 69.1 \\
     2 & 50 & Dark + Clean          & 57.2 & 67.2 & 69.7 & 69.7 \\
\midrule
     3 & 50 & Bright + Clean        & {\color{green}\textbf{62.5}} & 70.7 & 72.6 & {\color{green}\textbf{72.5}} \\
     3 & 50 & Bright + Dark + Clean & {\color{green}\textbf{57.5}} & 68.6 & 70.3 & {\color{green}\textbf{69.6}} \\
     3 & 50 & Dark + Clean          & {\color{green}\textbf{58.3}} & 72.2 & 72.5 & {\color{green}\textbf{71.5}} \\
\midrule
     4 & 25 & Bright + Clean        & {\color{red}\textbf{49.1}} & 69.7 & 71.1 & 66.5 \\
     4 & 25 & Bright + Dark + Clean & {\color{red}\textbf{59.1}} & 71.5 & 72.0 & 71.2 \\
     4 & 25 & Dark + Clean          & {\color{red}\textbf{50.4}} & 70.0 & 70.3 & 65.8 \\

\bottomrule
\end{tabular*}

\smallskip
\scriptsize
\begin{tablenotes}
\RaggedRight
\item[a] \(n_a\): No. of clients to go through adversarial training
\item[b] $ABR$: Adversarial training batch ratio
\item[c] $TDT$: Test Data Type
\item[d] $GA$: Global Accuracy
\end{tablenotes}
\end{threeparttable}
\end{table}
%%%%%%%%%%%%%%%%%%%%%%%%%%%%%%%%%%%%%%%%%%%%%%%%%%%%%%%%%%%%%%%%%%%%%%%%%%%
%%%%%%%%%%%%%%%%%%%%%%%%%%%%% Table VII %%%%%%%%%%%%%%%%%%%%%%%%%%%%%%%%%%%
\begin{table}[h]
\begin{threeparttable}
\caption{Robust Federated Learning (Non-IID) with two random clients consist of "bright" images. Evaluated on augmented test data}

\label{tab:2}
\setlength\tabcolsep{0pt} % make LaTeX figure out intercolumn spacing

\begin{tabular*}{\columnwidth}{@{\extracolsep{\fill}} ll ccccc}
\toprule
     \(n_a\)\tnote{a} & $ABR$\tnote{b}(\%) & $TDT$\tnote{c} & $GA$\tnote{d}(\%) &
     \multicolumn{3}{c} {Robust Acc.(\%)}  \\ 
     
\cmidrule{5-7}
    &&&& FGSM & FFGSM & Square\\

\midrule
     1 & 50 & Bright + Clean        & 55.0 & 58.1 & 60.4 & 60.8 \\
     1 & 50 & Bright + Dark + Clean & 47.8 & 66.1 & 66.3 & 64.5 \\
     1 & 50 & Dark + Clean          & 47.0 & 62.2 & 63.9 & 62.1 \\
\midrule
     2 & 50 & Bright + Clean        & 64.2 & {\color{green}\textbf{66.3}} & {\color{green}\textbf{66.9}} & {\color{green}\textbf{68.6}} \\
     2 & 50 & Bright + Dark + Clean & 54.3 & {\color{green}\textbf{68.4}} & {\color{green}\textbf{70.3}} & {\color{green}\textbf{68.2}} \\
     2 & 50 & Dark + Clean & 56.8 & 66.8 & 67.2 & 68.5 \\
\midrule
    3 & 50 & Bright + Clean        & 62.8 & 69.9 & 71.6 & 70.7 \\
    3 & 50 & Bright + Dark + Clean & 44.8 & 70.5 & 71.1 & 62.0 \\
    3 & 50 & Dark + Clean          & 49.2 & 68.2 & 68.3 & 66.4 \\
\midrule
    4 & 50 & Bright + Clean        & {\color{red}\textbf{38.6}} & 69.9 & 70.1 & 59.2 \\
    4 & 50 & Bright + Dark + Clean & 47.4 & 71.2 & 71.9 & 63.9 \\
    3 & 50 & Dark + Clean          & {\color{red}\textbf{36.1}} & 70.2 & 70.5 & 61.8 \\

\bottomrule
\end{tabular*}

\smallskip
\scriptsize
\begin{tablenotes}
\RaggedRight
\item[a] \(n_a\): No. of clients to go through adversarial training
\item[b] $ABR$: Adversarial training batch ratio
\item[c] $TDT$: Test Data Type
\item[d] $GA$: Global Accuracy
\end{tablenotes}
\end{threeparttable}
\end{table}
%%%%%%%%%%%%%%%%%%%%%%%%%%%%%%%%%%%%%%%%%%%%%%%%%%%%%%%%%%%%%%%%%%%%%%%%%%%

%%%%%%%%%%%%%%%%%%%%%%%%%%%%%%%%%%%%%%%%%%%%%%%%%%%%%%%%%%%%%%%%%%%%%%%%%%%
\begin{figure}
    \centering
    \subfigure[]{\includegraphics[width=0.24\textwidth]{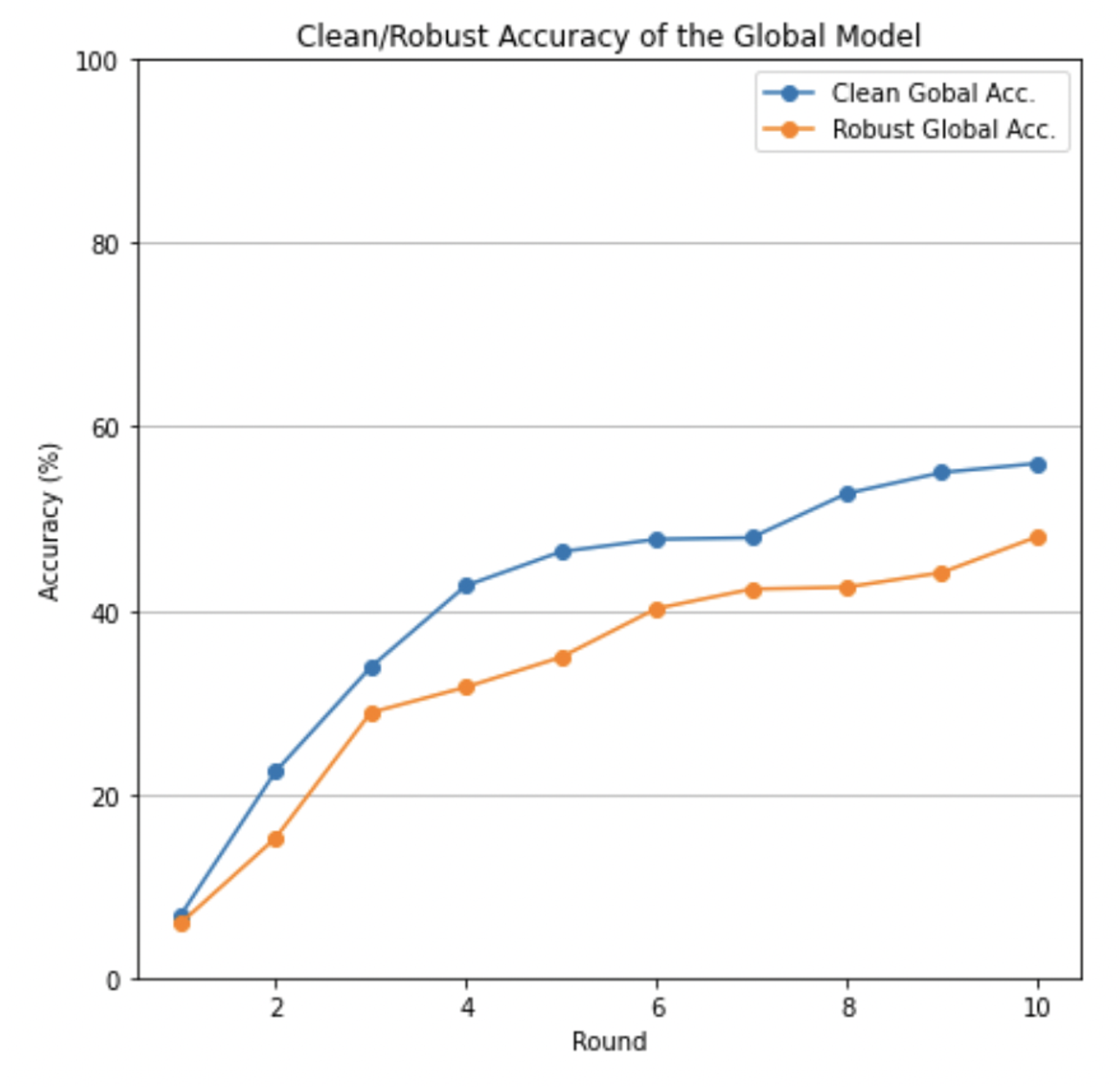}} 
    \subfigure[]{\includegraphics[width=0.24\textwidth]{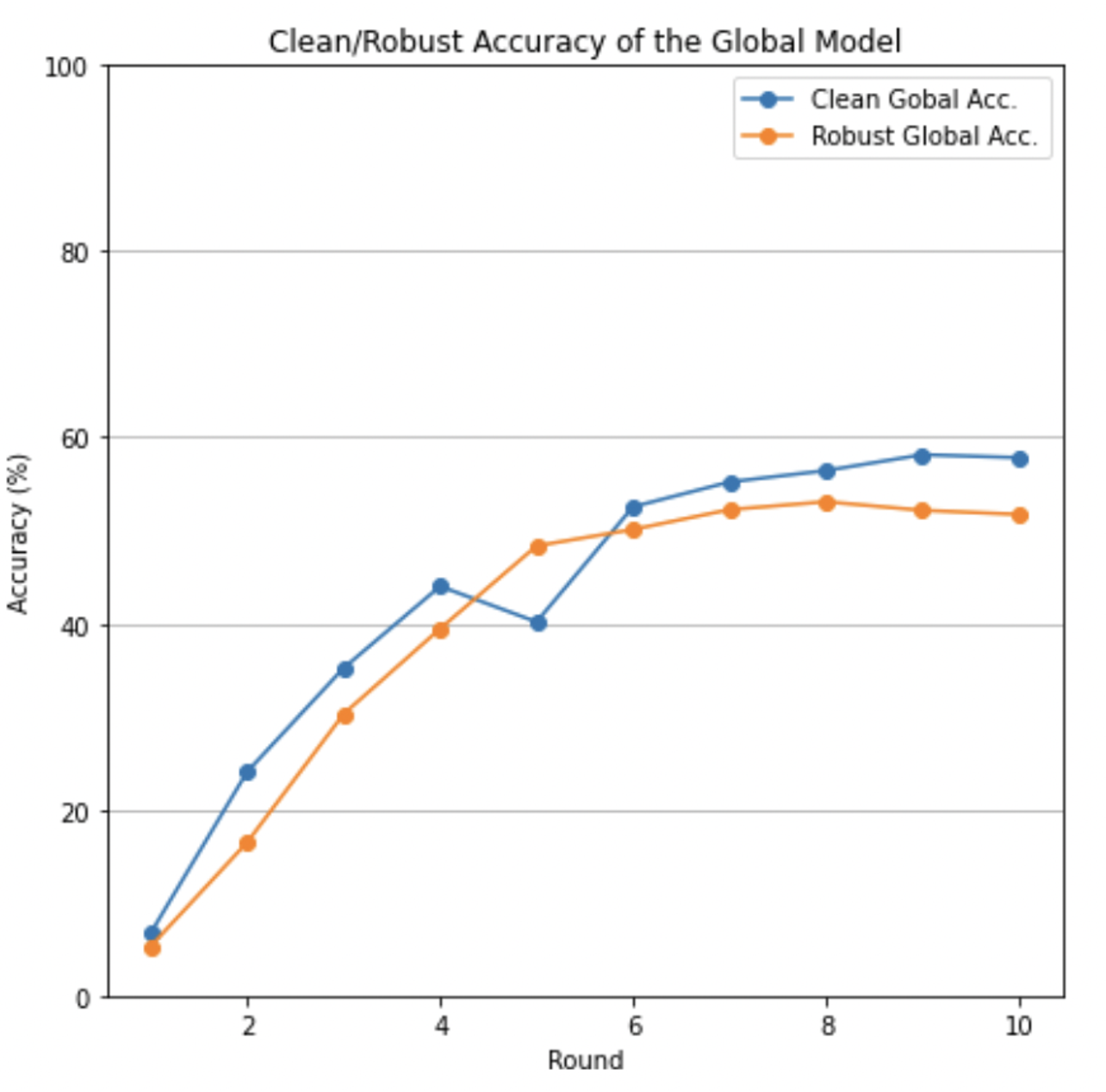}} 
    \subfigure[]{\includegraphics[width=0.24\textwidth]{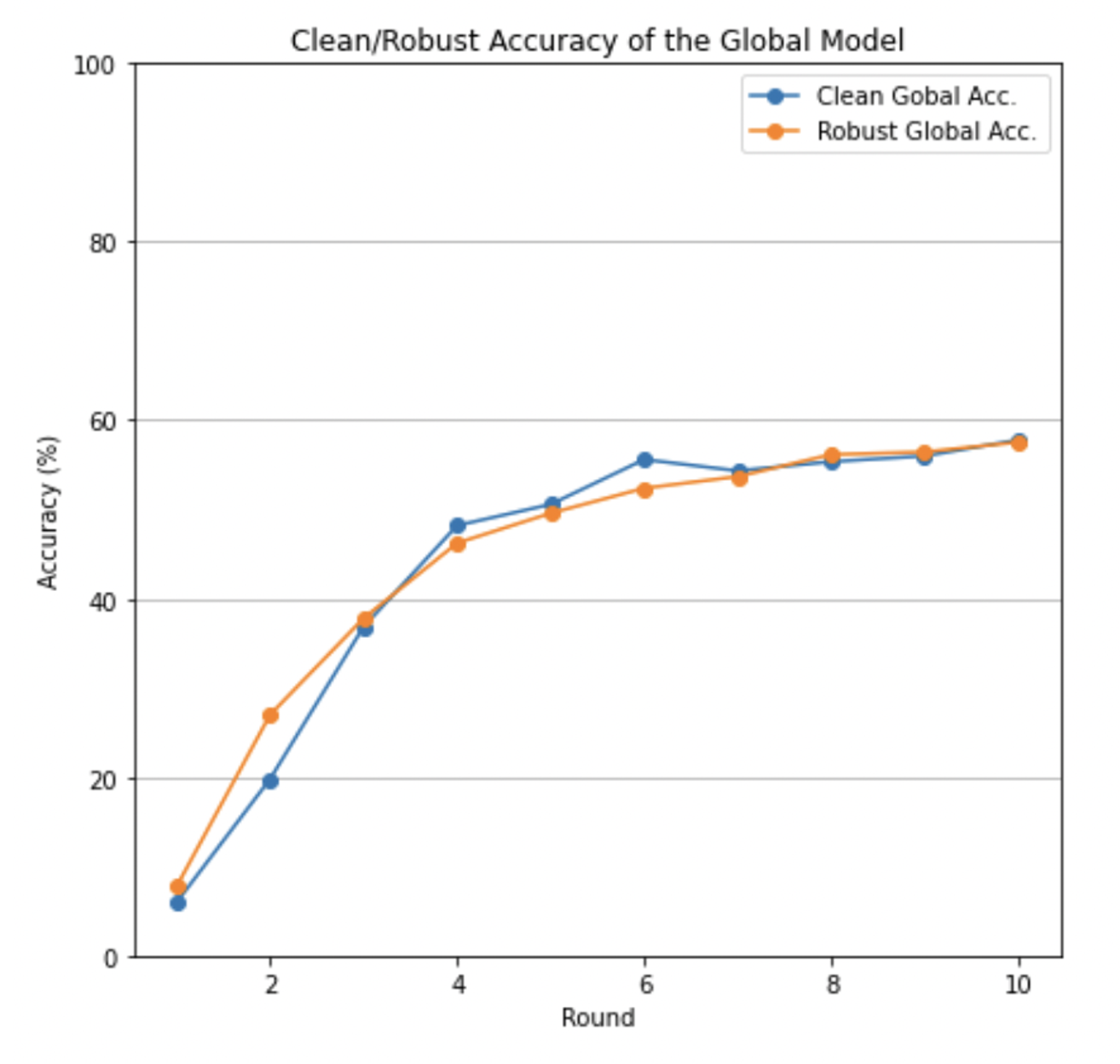}} 
    \subfigure[]{\includegraphics[width=0.24\textwidth]{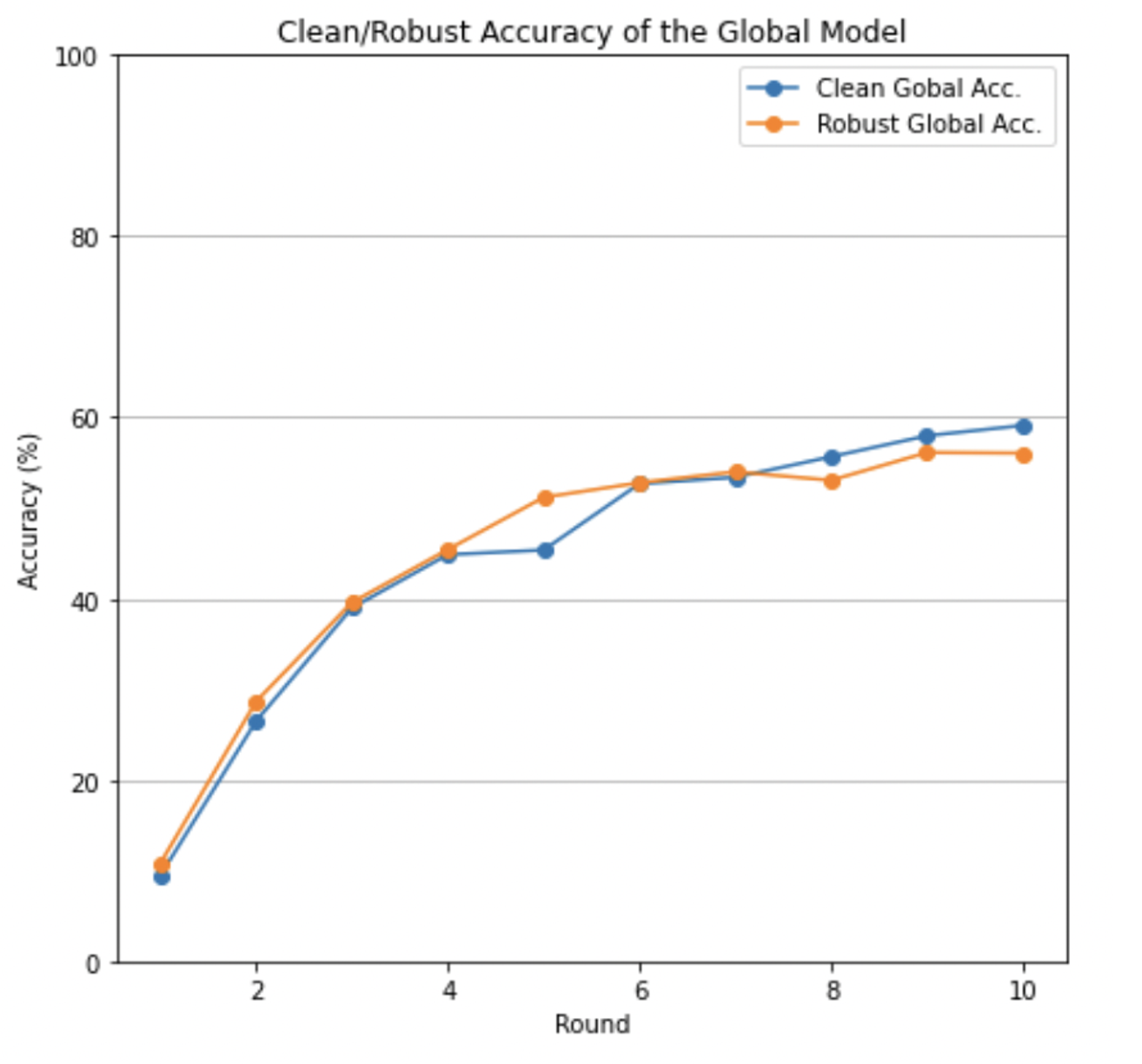}}
    \caption{Clean \& Robust Acc.(\%) of Robust Federated Learning (Two Random Clients with "Dark" Images). Evaluated on Augmented Test Data (\(TDT\): Bright + Dark + Clean). (a) \(n_a\)=1 (b) \(n_a\)=2 (c) \(n_a\)=3 (d) \(n_a\)=4}
    \label{fig:foobar}
\end{figure}
%%%%%%%%%%%%%%%%%%%%%%%%%%%%%%%%%%%%%%%%%%%%%%%%%%%%%%%%%%%%%%%%%%%%%%%%%%%

%%%%%%%%%%%%%%%%%%%%%%%%%%%%%%%%%%%%%%%%%%%%%%%%%%%%%%%%%%%%%%%%%%%%%%%%%%%
\begin{figure}
    \centering
    \subfigure[]{\includegraphics[width=0.24\textwidth]{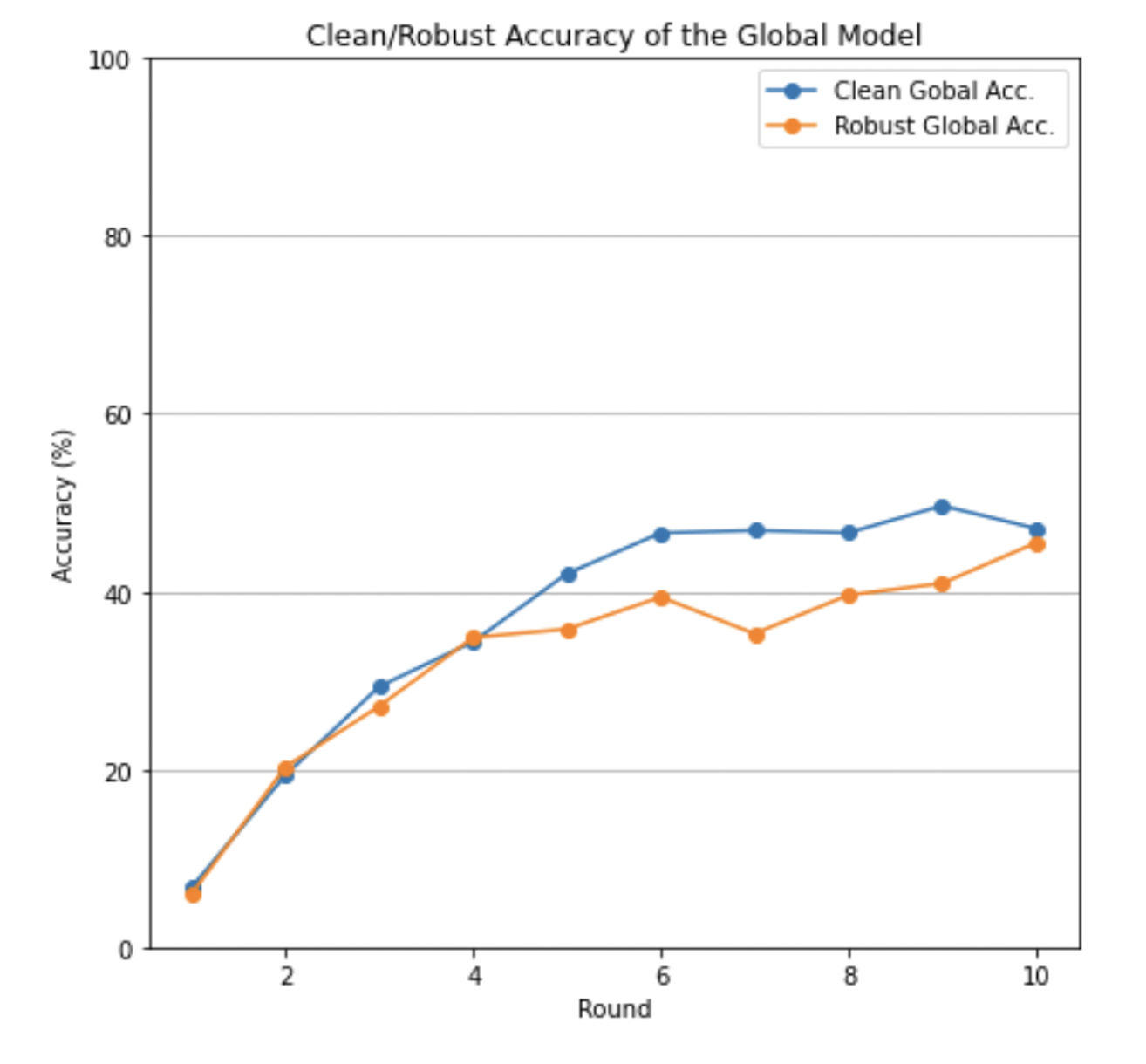}} 
    \subfigure[]{\includegraphics[width=0.24\textwidth]{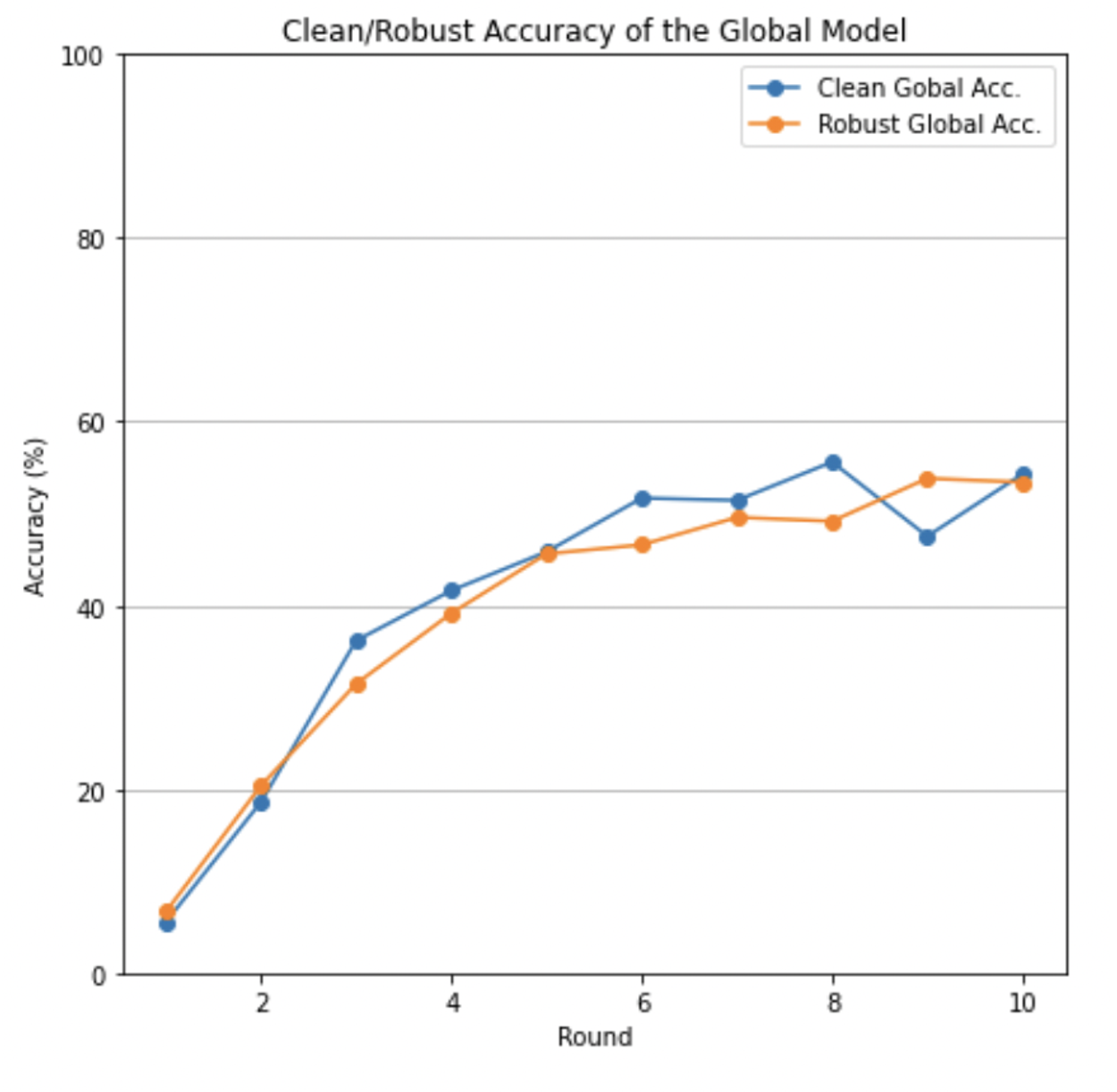}} 
    \subfigure[]{\includegraphics[width=0.24\textwidth]{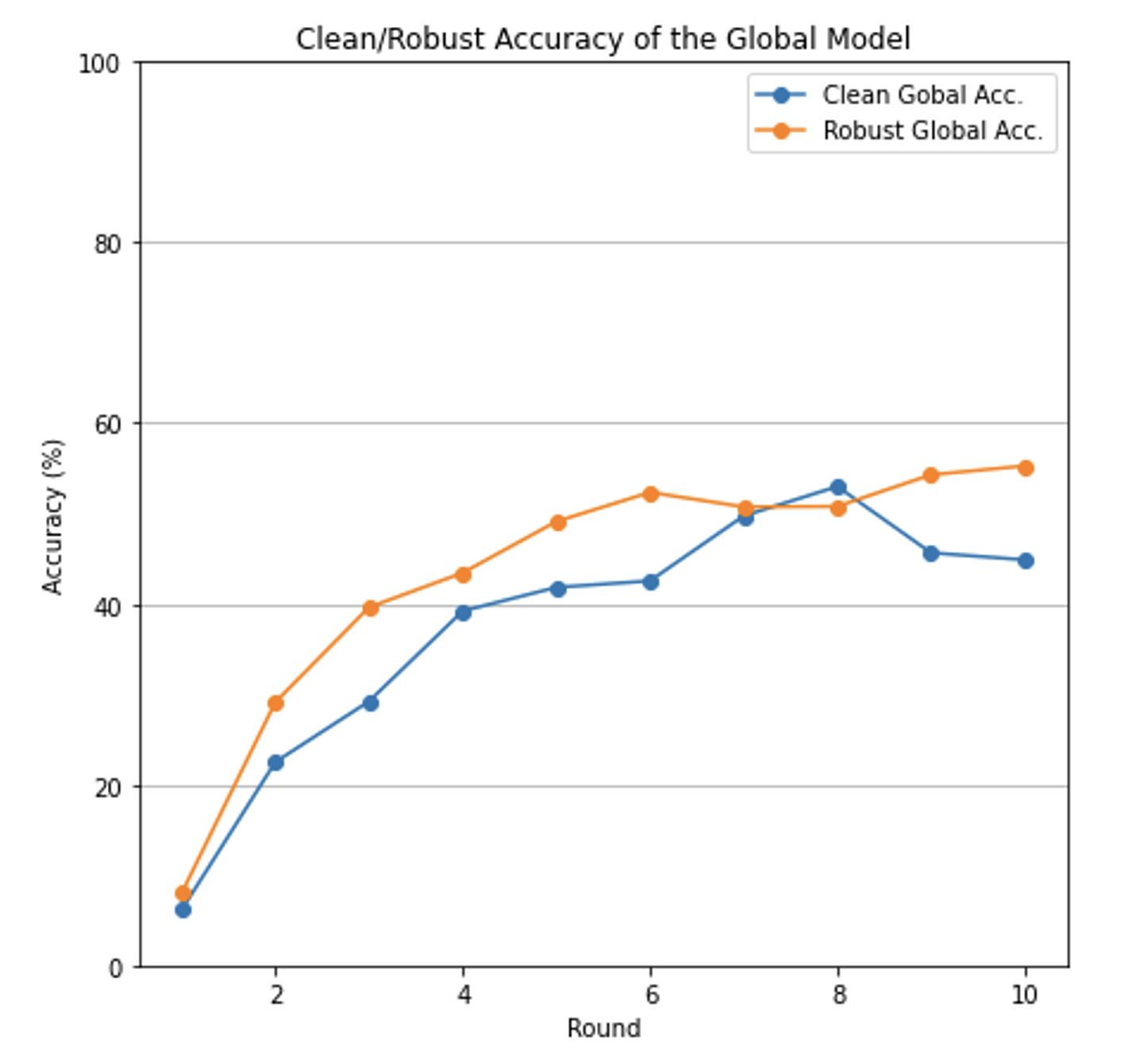}}
    \subfigure[]{\includegraphics[width=0.24\textwidth]{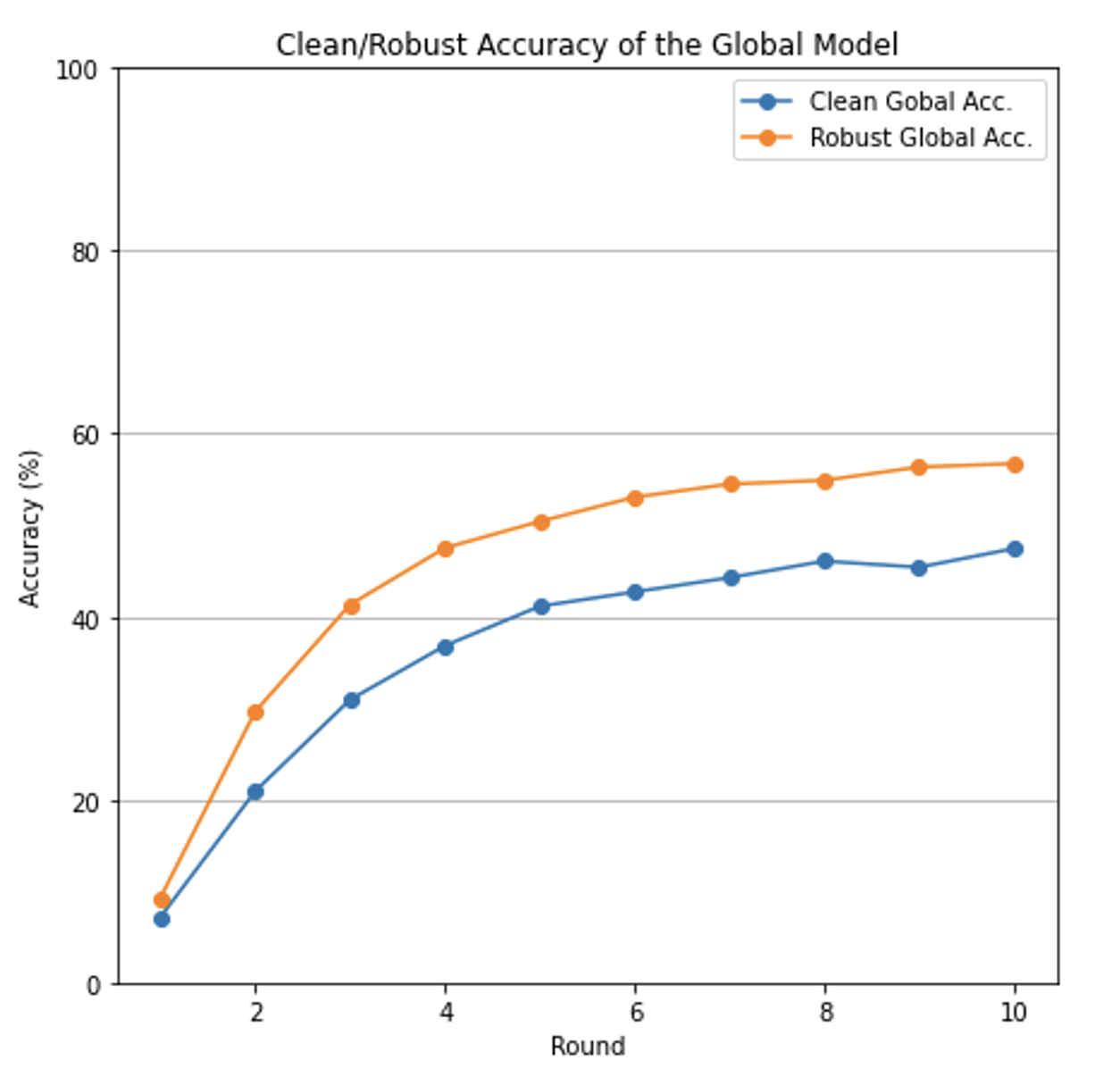}}
    \caption{Clean \& Robust Acc.(\%) of Robust Federated Learning (Two Random Clients with "Bright" Images). Evaluated on Augmented Test Data (\(TDT\): Bright + Dark + Clean). (a) \(n_a\)=1 (b) \(n_a\)=2 (c) \(n_a\)=3 (d) \(n_a\)=4}
    \label{fig:foobar}
\end{figure}
%%%%%%%%%%%%%%%%%%%%%%%%%%%%%%%%%%%%%%%%%%%%%%%%%%%%%%%%%%%%%%%%%%%%%%%%%%%

For previous experiments, the test dataset were only consisted of clean face images with no brightness modifications added. In this section, we evaluated our global model against different types of test dataset where we augmented brightness modified images. From Table VI, we may notice that most of the global accuracies were decreased from those from Table V. However, when \(n_a\) = 3, both global and  the robust accuracies against Square Attack \cite{andriushchenko2020square} achieved higher for all different \(TDT\). In addition, when we trained two randomly selected clients with bright images, robust accuracies of the global model against all three adversarial attack algorithms also increased when \(n_a\) = 2. From Fig. 11 and Fig. 12, we may further notice that the global accuracy trend has been stabilized with no notable fluctuations compared to those from Fig. 9 and Fig. 10, which is an positive signal which we must not ignore. 

%%%%%%%%%%%%%%%%%%%%%%%%%%%%%%%%%%%%%%%%%%%%%%%%%%%%%%%%%%%%%%%%%%%%%%%%%%%
\section{Discussion}
To begin with, one of the main limitations of our experiment is the variation of adversarial attacks. Although we have used three main methods to process adversarial training, both FGSM \cite{goodfellow2014explaining} and FFGSM \cite{wong2020fast} are within the family of fast adversarial example generation, which incorporate similar methods. The experiment would have been more compelling if other attack algorithms such as DeepFool \cite{moosavi2016deepfool}, One Pixel Attack \cite{su2019one} and Jitter-based Attack \cite{schwinn2021exploring} are used either to train the local models or to evaluate the global models. However, it was more time consuming to create adversarial examples with such methods especially when the face image data was set to 3 \(\times\) 224 \(\times\) 224. Lowering the pixel size will degrade both the clean and robust global accuracy, which hinders the main objective of our experiment. The adversarial attack methods that we chose were also able to demonstrate the robustness of our proposed federated learning algorithm and also the transferability of adversarial examples in a federated learning environment.

 Another restriction is that we have only used one weight averaging method, which could have limited the global model accuracy of our experiment. There are numerous other federated learning algorithms such as FedNova \cite{wang2020tackling}, FedRobust \cite{reisizadeh2020robust}, FedSVRG \cite{konevcny2016federated}, FedProx \cite{li2020federated}, etc. Such algorithms might have aggregated the parameters of each local models more constructively, hence achieving higher accuracy. 

Furthermore, we experimented with a starving federated data compared to other major face recognition benchmark datasets such as MegaFace Challenge \cite{kemelmacher2016megaface}, LFW \cite{srivastava2019performance}, and YTF, which our algorithm not reflect the SOTA face recognition ability of the trained model. The global accuracy could have been enhanced if we utilized such large datasets as well as other SOTA face recognition models such as ArcFace \cite{deng2019arcface}, CosFace \cite{wang2018cosface}, and FaceNet \cite{schroff2015facenet}. However, applying such datasets to the SOTA architectures would have been both computationally expensive and time consuming. Increasing the number of local devices for extended experiments will slow down the federated learning training process as well. As mentioned earlier, our investigation can be an illustration of a real-world scenario when some devices are locally trained with fairly smaller face image data size where we can observe how such experimental setting can influence the clean and robust global model accuracy.

Finally, the algorithm that we proposed may not directly correlate in solving the security issue of smart home face recognition system through federated learning. This is also because we did not implement the hardware aspect of the general security. However, since we tested with a total of 10 number of clients, each device contained around 2,454 face images, which is a more realistic amount of face data that a single smart home may possess. Moreover, the novelty in our empirical results have demonstrated that a smart home face recognition system can preserve its robustness even when the data is manipulated in the first place. But, further investigation with a bigger dataset and a increased number of total clients is needed to verify our results.

For further studies, we would like to propose researchers in this academic field to experiment with other domains. Robustness in Natural Language Processing (NLP) has been a challenge due to the discreteness of the text data. Any perturbations will always be apparent to our naked human eyes and modifying the vector space of the token will not be enough to make the language model robust enough. Recently, after the introduction of BadNets \cite{gu2017badnets}, backdoor attack in NLP has achieved high attack success rate in large datasets such as DBpedia \cite{socher2013recursive}, AG News \cite{zhang2015character}, and SST-2 \cite{socher2013recursive} by injecting backdoor trigger. Other attack methods such as paraphrasing the input text \cite{qi2021hidden} or replacing the keywords with their synonyms \cite{jin2020bert} has also led to a successful misclassification of well-known language models such as BERT \cite{devlin2018bert}. However, there are only two research conducted regarding backdoor defense in NLP (BKI \cite{chen2021mitigating} and ONION \cite{qi2020onion}). Since there are only handful of experiments processed in improving the robustness of a language model under a federated learning environment, exploration in this field of area will be intriguing. 

In addition, our investigation may have brought up the issue of bias of DNNs in face recognition models by modifying the brightness of the images. Recent works proposed by Yu et al.\cite{yu2021boosting} showed that an augmented dataset of clean and masked face images was able to improve the fairness of a single face recognition model. Further experiment in a starving robust federated environment and different types of data manipulation such as data augmentation and background color intensity modification may also improve the robustness of face recognition models utilized in smart homes.

\section{Conclusion}
Overall, we investigated the robustness of federated learning in smart home face recognition system. After applying number of different settings to the original benign federated learning architecture, we observed some interesting findings in regards to clean and robust accuracy of the global model. One of the major discovery from our experiment was when we included the brightness modified face images to the training process. Model lost its classification ability to the clean images where the standard global accuracy fluctuated chaotically. However, the aggregated model preserved its robust accuracy in an increasing trend against perturbed images generated based on several adversarial attack algorithms. 

In conclusion, we would like to emphasize the significance of this research in a way that adversarial attack will not only damage a single face recognition model embedded in smart homes, but it can also seriously destruct the security of an entire city that are empowered by the global model trained in a federated learning environment. The process of adversarial training must be a mandatory procedure included in the standard federated learning training so that any aggregated model distributed in the real-world will maintain its robustness against novel adversarial attack algorithms that are thoroughly researched in this field of area.

\bibliographystyle{IEEEtran}
\bibliography{custom}

\end{document}